\documentclass[lettersize,journal]{IEEEtran}
\usepackage{amsmath,amsfonts}
\usepackage{algorithmic}
\usepackage{algorithm}
\usepackage{array}
\usepackage[caption=false,font=normalsize,labelfont=sf,textfont=sf]{subfig} 

\usepackage{textcomp}
\usepackage{stfloats}
\usepackage{verbatim}
\usepackage{graphicx}
\usepackage{xcolor}
\usepackage{cite}
\usepackage[hyphens]{url}

\usepackage{background}
\usepackage{stackengine}

\usepackage[acronym]{glossaries}

\usepackage{multirow}
\usepackage{enumitem}

\newacronym{ecu}{ECU}{Electronic Control Unit}
\newacronym{bev}{BEV}{Battery Electric Vehicle}
\newacronym{phev}{PHEV}{Plug-In Hybrid Vehicle}
\newacronym{td}{TD}{Trip Distance}
\newacronym{ttl}{TTL}{Time To Leave}
\newacronym{qr}{QR}{Quantile Regression}
\newacronym{knn}{KNN}{K-Nearest Neighbours}
\newacronym{qknn}{QKNN}{Quantile K-Nearest Neighbours}
\newacronym{qarf}{QARF}{Quantile Adaptive Random Forest}
\newacronym{mcnn}{MCNN}{Feed-Forward Neural Network}
\newacronym{ffnn}{FFNN}{Feed-Forward Neural Network}
\newacronym{lstm}{LSTM}{Long Short-Term Memory}
\newacronym{mad}{MAD}{Mean Absolute Deviation}
\newacronym{mae}{MAE}{Mean Absolute Error}
\newacronym{mape}{MAPE}{Mean Absolute Percentage Error}
\newacronym{rmse}{RMSE}{Root Mean Squared Error}
\newacronym{mse}{MSE}{Mean Squared Error}
\newacronym{mpiw}{MPIW}{Mean Prediction Interval Width}
\newacronym{picp}{PICP}{Prediction Interval Coverage Probability}
\newacronym{adwin}{ADWIN}{Adaptive Windowing}
\newacronym{sgd}{SGD}{Stochastic Gradient Descent}
\newacronym{adam}{ADAM}{Adaptive Moment Estimation}
\newacronym{vif}{VIF}{Variance Inflation Factor}
\newacronym{soc}{SoC}{State of Charge}

\begin{document}

\title{
Online Learning Models for Vehicle Usage Prediction During COVID-19}

\author{Tobias~Lindroth, Axel~Svensson, Niklas~{\AA}kerblom, Mitra~Pourabdollah, and Morteza~{Haghir Chehreghani}
\thanks{Manuscript received October 28, 2022. This work was supported in part by Volvo Car Corporation and in part by the Department of Computer Science and Engineering, Chalmers University of Technology. The work of N.~{\AA}kerblom was supported in part by the Strategic Vehicle Research and Innovation Programme (FFI) of Sweden, through the project EENE (reference number: 2018-01937).}
\thanks{T.~Lindroth, A.~Svensson and M.~Pourabdollah are with Volvo Car Corporation, Torslanda, SE-405~31 G{\"o}teborg, Sweden (e-mail: tobias.lindroth@volvocars.com, axel.svensson.4@volvocars.com, mitra.pourabdollah@volvocars.com)}
\thanks{N.~{\AA}kerblom is with Volvo Car Corporation, Torslanda, SE-405~31, G{\"o}teborg, Sweden, and also with the Department of Computer Science and Engineering, Chalmers University of Technology, SE-412~96 G{\"o}teborg, Sweden (e-mail: niklas.akerblom@chalmers.se)}
\thanks{M.~{Haghir~Chehreghani} is with the Department of Computer Science and Engineering, Chalmers University of Technology, SE-412~96 G{\"o}teborg, Sweden (e-mail: morteza.chehreghani@chalmers.se)}}

\maketitle

\setstackEOL{\\}
\SetBgContents{\stackunder[37.5cm]{\Longstack{ This article has been accepted for publication in IEEE Transactions on Intelligent Transportation Systems. This is the author's version which has not been fully edited and\\ 
content may change prior to final publication. Citation information: DOI 10.1109/TITS.2024.3361676}}{$\copyright$ 2024 IEEE. Personal use is permitted, but republication/redistribution requires IEEE permission. See https://www.ieee.org/publications/rights/index.html for more information.}}
\SetBgScale{0.7}
\SetBgAngle{0}
\SetBgPosition{current page.center}
\SetBgVshift{0.5cm}
\SetBgColor{black}
\SetBgOpacity{1}

\begin{abstract}
Today, there is an ongoing transition to more sustainable transportation, for which an essential part is the switch from combustion engine vehicles to battery electric vehicles (BEVs). BEVs have many advantages from a sustainability perspective, but issues such as limited driving range and long recharge times slow down the transition from combustion engines. One way to mitigate these issues is by performing battery thermal preconditioning, which increases the energy efficiency of the battery. However, to optimally perform battery thermal preconditioning, the vehicle usage pattern needs to be known, i.e., how and when the vehicle will be used. This study attempts to predict the departure time and distance of the first drive each day using online machine learning models. The online machine learning models are trained and evaluated on historical driving data collected from a fleet of BEVs during the COVID-19 pandemic. Additionally, the prediction models are extended to quantify the uncertainty of their predictions, which can be used to decide whether the prediction should be used or dismissed. Based on our results, the best-performing prediction models yield an aggregated mean absolute error of 2.75 hours when predicting departure time and 13.37 km when predicting trip distance. 
\end{abstract}

\begin{IEEEkeywords}
Online machine learning, Uncertainty quantification, Vehicle usage prediction, COVID-19 pandemic.
\end{IEEEkeywords}

\section{Introduction}\label{introduction}
There is currently an ongoing transition to more sustainable transportation. Recently, there have been several initiatives that aim to reduce greenhouse gas emissions from vehicles and further accelerate the transition to zero-emission vehicles. The European Commission has, for example, proposed a 55 \% reduction of emissions from cars by 2030 and zero emissions from new cars by 2035 \cite{euroCommisionProposal}. Furthermore, several governments, cities, and automotive manufacturers have recently committed to work towards only allowing zero emission vehicles to be sold in the leading markets by 2035 and globally by 2040 \cite{govCOP26}. In California, an executive order has been issued requiring every new sale of passenger vehicles to be emission free by 2035 \cite{californiaExecutiveOrder}. 
An essential step in the switch to a more sustainable transportation system is battery electric vehicles (\acrshort{bev}s). \acrshort{bev}s are pure electric vehicles and make up two-thirds of the rapidly growing global electric car stock \cite{iea_2021}. While production of electric vehicles may produce emissions, usage is emission free as long as the energy sources used for charging do not emit greenhouse gases. 

However, even though \acrshort{bev}s have many advantages from a sustainability perspective, they face other challenges. Compared to vehicles with a combustion engine, \acrshort{bev}s have a more limited driving range and longer recharge times. These disadvantages are exacerbated in cold and hot climates due to harsher driving conditions, making propulsive consumption increase. Additionally, in these conditions, auxiliary consumption generally increases to keep satisfactory comfort when driving, especially in colder temperatures \cite{precond_paper, ev_consump_ambient_temp}.
There are several strategies to mitigate the effects caused by climate conditions, e.g., cabin preconditioning before departure, battery thermal preconditioning before fast charging, or battery thermal preconditioning before departure. The charging scheduling could be also optimized, as the time of day affects the cost and carbon dioxide footprint of the electricity production.

To perform all these different kinds of strategies, it is, however, needed to first learn about the vehicle usage pattern, i.e., how and when the vehicle will be used the next time. Some of this information, such as departure time and driving distance, can be given by the driver. However, it might not be convenient for the driver to fill in information about the next drive, and even if they do, it might not be accurate. Therefore, a better way to acquire the needed information may be to use historical data and machine learning to predict when and how the vehicle will be used in the near future. Consequently, this study explores how accurately different machine learning models can predict vehicle usage patterns. In particular, we attempt to predict departure times and driving distances. 

The data used in this study arrive over time, and as argued in \cite{LOSING20181261}, a specific type of machine learning, called \emph{online learning}, fits well with this setting. Online learning is a machine learning paradigm that continuously updates the models as new data arrive over time, and subsequently discards the data after the model has been updated. This way, the models may function in environments with limited data storage and computational power. These characteristics could allow us to implement the prediction models on the \acrshort{bev}s directly, avoiding the need to transfer sensitive user data over the network.

In the literature, there seems to be a lack of consensus on the precise definition of online learning, and it is often intertwined with a similar paradigm called incremental learning \cite{Gepperth2016, mukherjee_datta, NIPS2006_a92c274b, LOSING20181261}. To make these concepts clear, in this paper we refer to online learning as the concept of training sequentially on a single observation or a limited sequence of past observations.

The main contributions of this paper are summarized as follows:
\begin{itemize}[leftmargin=*]
    \item To the best of our knowledge, no research has been conducted on predicting vehicle usage using online learning models. Thus, this paper introduces a new perspective by utilizing novel and efficient prediction methods which can be used in computationally constrained environments, such as in the internal electronic control units (\acrshort{ecu}s) of a vehicle.
    \item This work utilizes novel approaches to estimate the quality of the predictions, which could enable the vehicle to invalidate uncertain predictions without any human intervention. 
    \item We introduce a new approach to estimate whether the vehicle has recurrent behavior using the clustering tendency of the departure times and driving distances.
    \item This work also considers the prediction of driving behavior during the COVID-19 pandemic, where the studied non-public real-world data set covers more than 300 \acrshort{bev}s driven over the course of an entire year. To our knowledge, no work has performed such a comprehensive analysis of predicting vehicle usage during COVID-19, where commutes are increasingly irregular.
\end{itemize}

The rest of this paper is organized as follows: 
Section \ref{related_work} covers related work. 
Section \ref{sec:problem_statement} concretizes the problem to be solved.
Section \ref{sec:data} describes the data and the preprocessing steps. In Section \ref{sec:prediction_process}, the prediction models are described together with the validation procedure. Section \ref{sec:results} covers the results for predicting driving distance and departure time, 
and, finally, Section \ref{sec:conclusion} concludes the paper.

\section{Related Work}\label{related_work}
This section describes the related work wherein departure time and trip distance forecasts are performed using various machine learning models. However, no existing work considers online learning models, and to the best of our knowledge, no research has been performed regarding the prediction of vehicle usage using online models.

\subsection{Departure Time Prediction} \label{Departure_time_forecast}
Intuitively, the forecast for individual \acrshort{bev}s should depend highly on the driver, compared to, e.g., busses that follow a fixed schedule. Considering an office commute, where flexible working hours are often applied, the arrival time to the office could vary significantly, which implicitly should affect the departure time from the office as well. This type of correlation is shown in \cite{departure_smart_charging}, where the authors predict the departure times from workplaces, to optimize smart charging. The authors point out that the average departure time of a \acrshort{bev} has a standard deviation of 141 minutes, showcasing the difficulty of this task. They investigate several different regression models on historical charging data and observe that sophisticated algorithms like XGBoost and neural networks perform better than linear models, receiving a mean absolute error of 82 minutes. In \cite{data_driven_approach_charing_behavior}, the authors predict the starting time and end time of an upcoming trip based on the start time, end time, and distance of the most recent previous trip. They receive a root mean square error (\acrshort{rmse}) of around 2 hours for both the start and end time. 
Other approaches to departure forecasting predict time intervals instead of a specific time. For example, \cite{zurich} forecasts a time interval of 15 minutes, resulting in an \acrshort{rmse} of roughly 3 hours. 
The authors of \cite{driving_behavior} investigate the variation of departure times and attempt to model the first daily departure time as a probability distribution. They state that this is impossible after applying statistical tests measuring the similarity of well-known probability distributions.

\subsection{Trip Distance Prediction} \label{travel_duration}
Trip distance is an important factor to know certain characteristics of the trip, and can be useful when, e.g., deciding whether battery thermal preconditioning is beneficial or not. Panahi et al. \cite{forecast_load_ann} predict the driving distance using neural networks, resulting in a mean absolute deviation (\acrshort{mad}) of around 9\%. Baghali et al. \cite{data_driven_approach_charing_behavior} implement a number of machine learning models such as \acrfull{knn}, decision trees, random forest, and neural networks for predicting the daily travel and charging demand of electric vehicles. The novelty in their work is that they consider charging at more places than at home. They show that even the less complex models generate reasonable results, with a \acrshort{rmse} of around 19-23 km. However, they point out that to find daily temporal patterns, more complex models are needed. 
As the authors of \cite{driving_behavior} state, different individuals may have different driving patterns. The authors of \cite{PEV_behavior_energy_market} cluster plug-in electric vehicle owners according to driving behaviors. They train two \acrfull{lstm} recurrent neural networks on each cluster to forecast arrival time and travel distance, and subsequently investigate the financial impacts of charging demand using these predictions. They show a significant improvement in forecasting the travel distance when the driving behavior is tied to a specific travel pattern.

\subsection{Driving Patterns During COVID-19}
The COVID-19 pandemic has had major impacts on people's everyday lives and how they travel. Not only due to the severity of the disease, but also how authorities enforced restrictions to counter the spread of the pandemic.  Interestingly, \cite{asi5010012} and \cite{9621740} found out that the vehicle charging behavior was affected by the level of restrictions enforced, where the charging activity increased as restrictions were lifted. 
To deal with the restrictions, various workplaces started to opt for teleworking \cite{RePEc:ema:worpap:2021-06, DEPALMA2022372, HiseliusWinslott, GaoLevisson, JAVADINASR2022466}. In fact, a survey in Canada shows that the amount of people working remotely increased five-fold \cite{RePEc:ema:worpap:2021-06, DEPALMA2022372}, while the number of people who commuted decreased slightly compared to the pre-pandemic period. A survey of public agencies in Sweden found that people who commuted 4-5 times a week pre-pandemic barely commuted once a week during the pandemic, due to teleworking \cite{HiseliusWinslott}. Further, it was reported the 86\% of the people changed commuting behavior during the pandemic, despite the relatively liberal restrictions in Sweden. In a large study of travel behavior, \cite{JAVADINASR2022466} included a survey of how people expected to commute in the future. It was reported that 48\% anticipate having the option to work from home, and that it is expected that people will commute 3.42 times per week, showing 17\% decline from the pre-pandemic period. 
The decrease in commuting trips has paved the path for new behaviors. \cite{RePEc:ema:worpap:2021-06, DEPALMA2022372} found that despite the decrease in commuting, the total number of trips increased. One reason for this might be that the time employees save by not commuting is used to perform other kinds of trips \cite{CALEARO2021111518}, \cite{HiseliusWinslott}. 

\section{Problem Statement}\label{sec:problem_statement}

\begin{figure}
    \centering
    \includegraphics[width=0.35\textwidth]{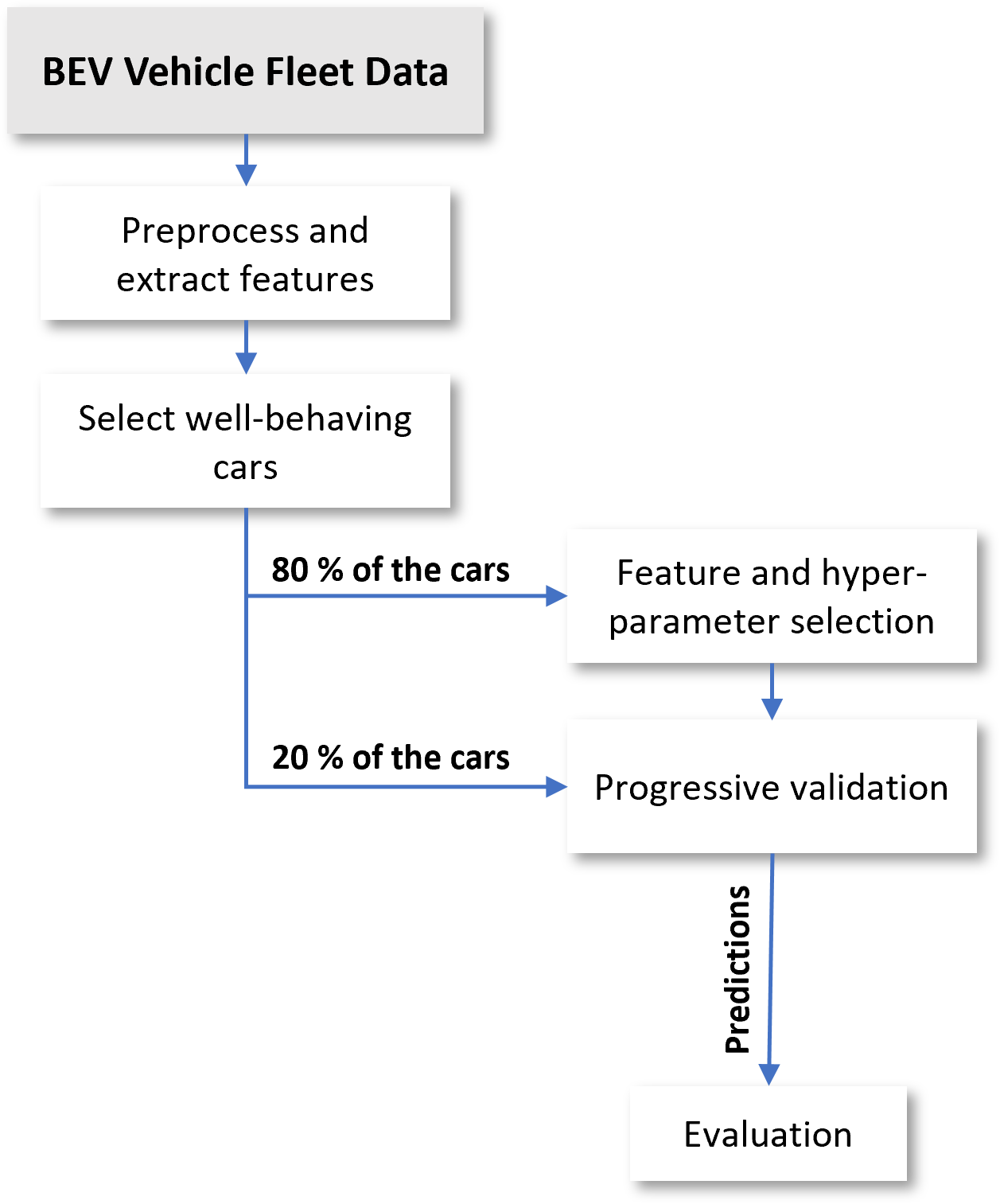}
    \caption{{Overview of our process for online learning model evaluation.}}
    \label{fig:overview}
\end{figure}

{In this paper, we study the question of \emph{how accurately online learning models can predict the departure time and distance of an upcoming drive}}.
We address this question by comparing several online machine learning models in terms of error rate, as well as how uncertain the models are in their predictions. The uncertainty is an important aspect since it can be essential to know how confident the model is when a prediction is used to make a critical decision. {The process we use to perform and evaluate the models is outlined in Figure \ref{fig:overview}. First, we identify a set of informative base features, which we then expand using feature engineering. Second, we select the subset of 100 cars from the original data set which we determine to be the most \textit{well-behaving}, i.e., cars for which there are identifiable patterns in the data, despite the pandemic. The resulting set of cars is split into training and test sets, the first of which is also used to select the best features and hyperparameter values for each model. Finally, we utilize progressive validation to compare and evaluate the performance of the models.}

The core online learning problem is defined as follows: 
For each time step $t$, where $t$ corresponds to a day, a feature vector $\textbf{x}_t$ is received by a prediction model $M_{t}$ at midnight. The model $M_{t}$ is expected to predict the target $y_t$, which can be either the time from midnight to the first departure time, or the driving distance of the first drive. When the true value of the target $y_t$ becomes available, the task is to infer a model $M_{t+1}$ only based on $(\textbf{x}_t, y_t)$ and $M_{t}$. The model $M_{t+1}$ is then used in the next time step. {Since we generate a large number of features and different sets of features are selected for each model, due to limited space, we only report the process used to select the features (in Section \ref{sec:data}), and the length of the feature vector $\textbf{x}_t$ used by each model (in Section \ref{sec:prediction_process}).}

To limit the scope of the study, the focus is put on commuting drives. For this reason, we only consider the first drive of each day for predictions. It is assumed that each prediction is performed at midnight and that it is known whether there will be a drive or not on that specific day. Further, the predictions are performed individually for each vehicle, i.e., each vehicle has its own prediction model. Online learning has, as mentioned earlier, the benefit of potentially being implemented in the \acrshort{ecu}s of the vehicles, as it is able to operate in a system with limited computational resources.  
{We emphasize that each model studied in this work has the essential property that the run-time and space complexity (during prediction, or training using a single new observation) is constant with respect to the number of previous observations.}

\section{Data}\label{sec:data}
The data used to train and evaluate the prediction models are collected from a fleet of company car \acrshort{bev}s, where each employee has signed an agreement allowing use of the data for research purposes. No personal or sensitive data have been accessed or used for this study. The data consist of pseudonymised measurements of different attributes such as velocity, acceleration, the state of charge (\acrshort{soc}), and energy consumption. {These attributes are measured at a constant rate of 0.1 Hz during measurement sessions and give a granular view of the car's properties.} From these measurements, we extract summaries of the driving and charging sessions, and use them as input to the prediction models. These summaries consist of aggregations of the attributes, such as mean, standard deviation, maximum and minimum values. 

\begin{figure}%
    \centering
    \captionsetup[subfloat]{font=scriptsize,labelfont=scriptsize}
    \subfloat[\centering Driving distance \label{fig:data_analysis_td}] {\includegraphics[width=0.4\textwidth]{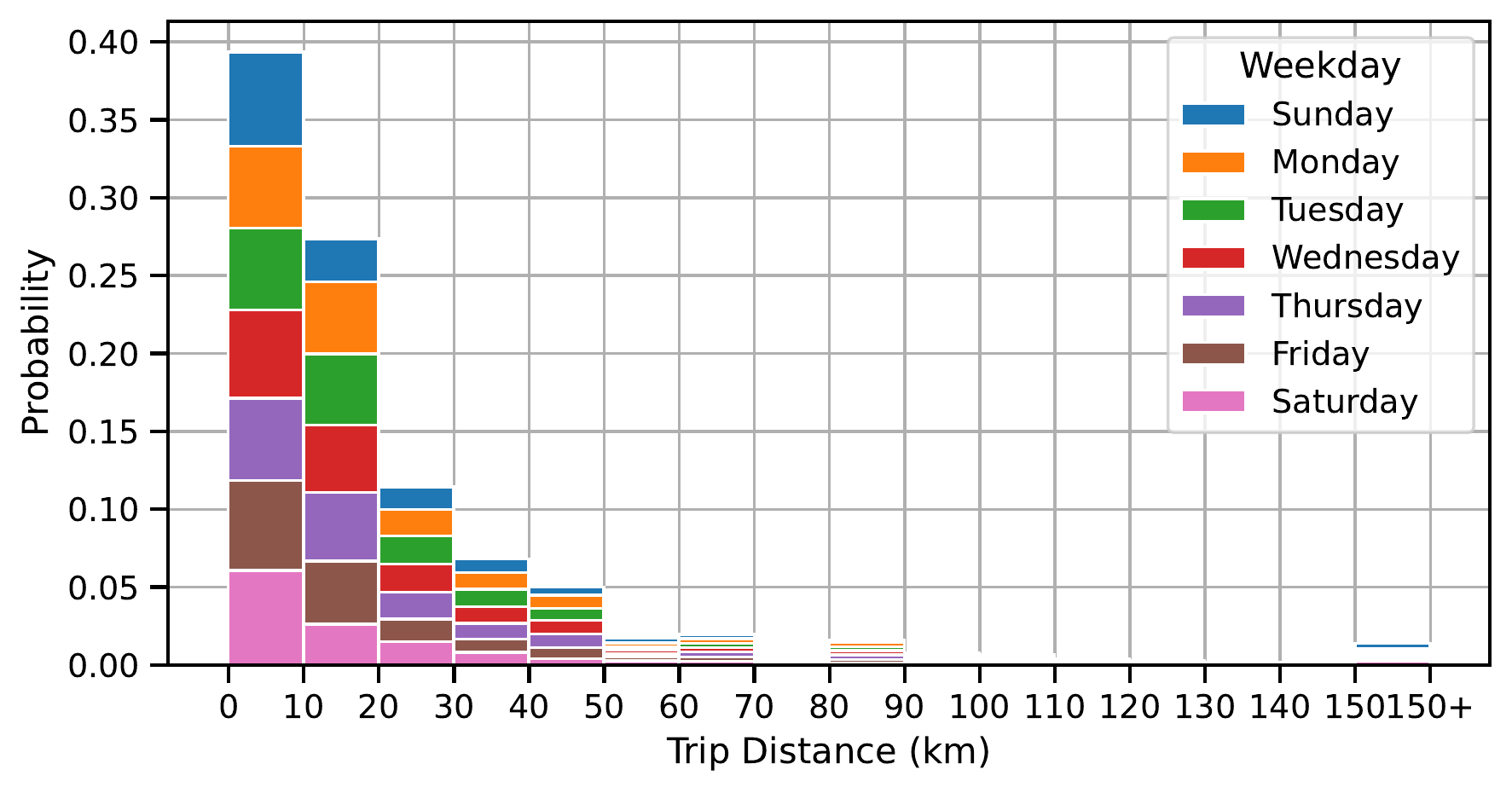}} \\
    \subfloat[\centering Departure time \label{fig:data_analysis_ttl}] {\includegraphics[width=0.4\textwidth]{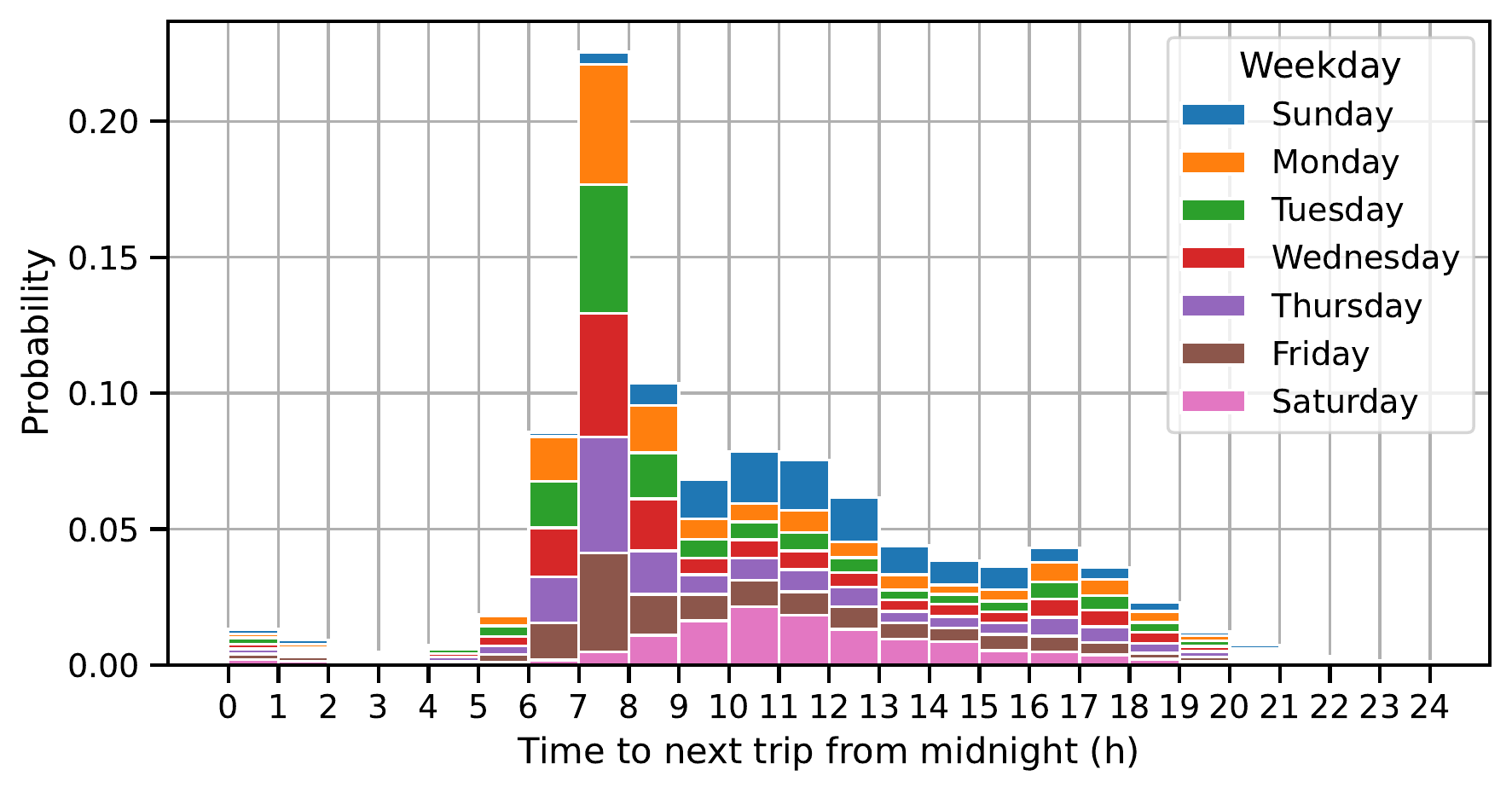}}
    \caption{{Distribution of trips in BEV data set, with respect to driving distance and departure time.}}
    \label{fig:data_analysis}
\end{figure}

{To briefly describe the trends of the variables we intend to predict, it can be seen in Figure \ref{fig:data_analysis_td} that most of the trips in the data set are short, with a mean distance of 23.1 km and a median distance of 13.3 km. Generally, each of the trip length intervals are equally likely all days of the week, except trips of length 10-20 km, which are slightly less common during weekends. In contrast, the few long-distance trips (150+ km) in the data more commonly occur during weekends. In Figure \ref{fig:data_analysis_ttl}, the departure time varies more than the trip distances, with a peak at roughly 7.5 a.m., and a mean departure time at 10.42 p.m. The peak only occurs between Monday and Friday, being entirely absent during weekends. Much lower peaks can be observed between 10 a.m. and 12 p.m., as well as around 16-17 p.m., where differences between weekends and other days are also significantly less pronounced.}

\subsection{Preprocessing}
To increase the data quality, we apply several filters to the summarizations of driving and charging sessions. These filters are described in the list below and handle inconsistencies such as missing data and abnormally short or long sessions.

\begin{itemize}[leftmargin=*]
    \item All driving and charging sessions shorter than 50 seconds are removed. The selection of 50 seconds is arbitrary, though it removes most of the anomalous short trips and charging sessions.
    \item Two subsequent drive or charging sessions are merged if there are less than 15 minutes between them. This mainly affects the driving sessions, and we argue that this provides a better overview of the vehicle usage and that a 15-minute break should not break one drive into two. 
    \item Vehicles with less than 50 drives are removed from the data to ensure there exists a minimum of drives for each vehicle from which the prediction models can attempt to learn.
\end{itemize}

\subsection{Features considered} \label{features}
As mentioned previously, the predictions of departure time and trip distance are based on aggregations of different signals: 

{\begin{itemize}[leftmargin=*]
    \item \textit{Date of when prediction is performed} 
    \item \textit{Start time, end time, and distance of the previous trip} 
    \item \textit{Start time of most recent charging session} 
    \item \textit{State of charge at the beginning of most recent charging session} 
    \item \textit{Ambient temperature and sun load during the previous trip} 
    \item \textit{Speed and acceleration during the previous trip} 
    \item \textit{State of charge during the previous trip} 
\end{itemize}}

Beyond these features, we perform feature engineering to extract further information from the temporal features and the previous target variables. Starting with the temporal features, we extract the minute of the hour, the hour of the day, the part of the day (morning, noon, afternoon, etc.), the day of the month, the day of the week, whether it is a workday or not, similar to \cite{departure_smart_charging}. The features that are categorical by nature, e.g., part of day, are described by one-hot encoding. Further, cyclic representations are extracted from some of the continuous temporal variables. This is done by performing the trigonometric transformations
\begin{align}
    f_x = \sin(2\pi f/\max(f)), \\
    f_y = \cos(2\pi f/\max(f)),
\end{align}

where $f$ is the feature to be transformed and $f_x$ and $f_y$ are the two components of the cyclic feature \cite{predict_ev_charging}. Lastly, inspired by \cite{predict_ev_charging, departure_smart_charging}, historical averages and running averages of the past target variables are introduced.

\section{Prediction Models}\label{sec:models}
In this section, we introduce the online machine learning models that we adapt to predict the departure time and distance of an upcoming drive. Furthermore, we compute the uncertainty of the predictions as well, to investigate the effects on changing user behavior.
To quantify the uncertainty of a prediction, one may use prediction intervals. A prediction interval quantifies a potential range where a future observation will occur with a certain probability. For instance, when using a 95\% prediction interval, we expect that the next observation will fall within the prediction interval with a 95\% certainty. 

A prediction interval may be defined as 
\begin{equation}
    [\hat{y}-z_{\frac{\tau}{2}}\sigma, \hat{y}+z_{\frac{\tau}{2}}\sigma] ,
    \label{eq:pred_error}
\end{equation}
where $\hat{y}$ is the model prediction, $z_{\frac{\tau}{2}}$ is the upper $\frac{\tau}{2}$ quantile of a standard normal distribution and $\sigma$ is the prediction standard error \cite{zhu2017deep}.

\subsection{Historical Average}
As a baseline, we use the historical average to predict the departure time and trip distance. This method is incremental by nature, as it keeps track of the average departure time and trip distance over time.

\subsection{Quantile Regression (QR)}
The first machine learning model we study is an online version of the quantile linear regression model. Ordinary linear regression is defined as
\begin{equation}
    \hat{y} = \theta^T \textbf{x} ,
    \label{eq:linearRegression}
\end{equation}
where $\hat{y}$ is the predicted target, $\textbf{x}$ is the observed feature vector and $\theta$ is the weight vector. The weight vector may be updated incrementally using stochastic gradient descent (\acrshort{sgd}) in the following way: 
\begin{equation}
    \theta^{(i+1)} = \theta^{(i)} - \eta\nabla(\mathcal{L}(y, \textbf{x}, \theta^{(i)}) + (\theta^{(i)})^T (\theta^{(i)}) \lambda) ,
    \label{eq:sgd}
\end{equation}
where $\lambda$ is a regularization parameter, {$\eta$ is a learning rate parameter}, and $\mathcal{L}(y, \textbf{x}, \theta^{(i)})$ is the loss function with the true target $y$, the observed feature vector $\textbf{x}$ and the current weight vector $\theta^{(i)}$. {The hyperparameters $\lambda$ and $\eta$ are tuned using grid search.} Using the squared loss, the predicted target will be the conditional mean. However, by changing the loss to the tilted absolute value function, we may predict a specified quantile instead. The tilted absolute value function is defined as
\begin{equation}
    L_\tau(y, \textbf{x}, \theta) = 
    \begin{cases} 
        (\tau-1)(y-\theta^T \textbf{x}) & \textrm{ if } y < \theta^T \textbf{x},\\
        \tau(y-\theta^T \textbf{x}) & \textrm{ if } y \geq \theta^T \textbf{x},
    \end{cases}
\end{equation}
where $\tau$ is the chosen quantile \cite{wang2022comprehensive, koenker2001quantile}. Using this function, the inference is divided into two parts. The inferred prediction by the model is the median, i.e., the $50^{\textrm{th}}$ quantile, while the uncertainty is defined by predicting the $5^{\textrm{th}}$ and the $95^{\textrm{th}}$ quantiles using the tilted absolute value function, resulting in $\hat{y}-z_{\frac{\tau}{2}}\sigma$ and $\hat{y}+z_{\frac{\tau}{2}}\sigma$ respectively in Equation \ref{eq:pred_error}. 

\subsection{Quantile K-Nearest Neighbours (QKNN)}
The second machine learning model is an adjusted version of the K-Nearest Neighbours (\acrshort{knn}) method. A prediction $\hat{y}$ is calculated by taking the mean of the closest $K$ neighbors:
\begin{equation}
    \hat{y} = \frac{1}{K} \sum_{\textbf{x}_i \in N_k(\textbf{x})} y_i ,
\end{equation}
where $N_k(\textbf{x})$ is the set of seen observations closest to the new observation, and $y_i$ is the target value for observation $\textbf{x}_i$ in the set $N_k(\textbf{x})$. A typical way of measuring the distance between observations is to use the Euclidean distance in the feature space \cite[pp.~14--16]{hastie2009elements}. The KNN model is made online by only using the $N$ latest observations, where $N$ is a tunable hyperparameter. {We tune both $N$ and $K$ using grid search.}
The quantiles in Equation \ref{eq:pred_error} are estimated by fitting a Gaussian distribution on the $K$ nearest neighbours and using the standard deviation of that distribution in Equation \ref{eq:pred_error} \cite{vasseur2021comparing}.

\subsection{Quantile Adaptive Random Forest (QARF)}
The third machine learning model is a modified version of Random Forest \cite[pp.~305--307]{hastie2009elements}. To make it online, we use incremental Hoeffding trees as base-learners \cite{Gomes2017, Gomes2018}. Hoeffding trees are described in \cite{Domingos2000}, and can determine, with a certain confidence, the number of observations needed to select the optimal splitting feature. In our setting, the feature to split at a specific node is the one reducing the variance in the target space the most \cite{FIMT-DD, streaming_forecasting}. Furthermore, to make the online Random Forest more responsive to changes in the data, we use an algorithm called the Adaptive Windowing to retrain a new version of the model in the background if a potential shift in data distribution is detected. {We treat the number of trees as a hyperparameter and tune it using grid search.}

The online Random Forest is then modified further for uncertainty estimation, using the method of Vasiloudis et al. \cite{vasiloudis2019quantifying}. Their approach is to keep an approximated representation of the observed target values, from which it is possible to extract percentiles at desired significance levels. To bound the memory required for storage of the approximation, the authors use an incremental and mergeable data structure called a \textit{KLL sketch}. They let each leaf in the forest store a sketch, which keeps an approximation of the target values with a small memory footprint. When the forest receives a new training instance, each tree sorts the instance to a leaf, and updates the leaf's sketch with the target of the training instance. The prediction process makes use of the approximated target values to create a prediction interval. Each tree sorts the input to a leaf, and all the sketches of the reached leafs are merged. From the merged sketch we fit a Gaussian distribution, and use the standard deviation of this distribution in Equation \ref{eq:pred_error}.

\subsection{Feed-Forward Neural Network with Uncertainty Quantification (MCNN)}
The last machine learning model we study is a Feed-Forward Neural Network, where each hidden layer $h$ is defined by 
\begin{equation}
    h_{i+1} = g(W^T h_i + b)
\end{equation}
where $W$ and $b$ are learnable parameters and $g(\cdot)$ is a non-linear activation function. This method is made online by only using a single observation at a time when performing back propagation to train the network, hence, training it sequentially as new observations are observed. {Using grid search, we tune the learning rate, the number of hidden layers, the number of neurons in each layer, and the dropout rate.}

When quantifying uncertainty, it is common to differentiate between two different kinds of uncertainty: \textit{aleatoric} and \textit{epistemic} \cite[pp.~7--9]{Gal2016UncertaintyID}, \cite{tagasovska2019single, zhu2017deep, lai2021exploring, KWON2020106816, shaker2020aleatoric}. Aleatoric uncertainty corresponds to the uncertainty present in the data, such as noise. However, it could also be due to hidden variables that affect the prediction outcome (in the context of our work, a hidden variable could, for example, be information about upcoming meetings at work). On the other hand, epistemic uncertainty corresponds to the uncertainty due to the prediction model being inexperienced in certain regions of the data distribution.  
To quantify the uncertainty in each prediction, we take inspiration from \cite{zhu2017deep}, who aggregate the epistemic and aleatoric uncertainties to estimate the standard error used in Equation \ref{eq:pred_error}. The epistemic uncertainty is estimated using a technique called Monte Carlo Dropout \cite{gal2016dropout}. In short, multiple predictions are performed using the same observation, where each prediction is individually affected by dropout. This means that some of the neurons in each layer are disregarded for each prediction, resulting in varying outputs. From a Bayesian perspective, the outputs can be seen as samples from the posterior predictive distribution. This posterior estimation approximates a Gaussian process, where the variance of the process corresponds to the uncertainty, approximated as
\begin{equation}
    \hat{\sigma}^2_e = \frac{1}{B} \sum_{b=1}^{B} (\hat{y}_b - \Bar{\hat{y}})^2 ,
\end{equation}
where $B$ is the number of predictions, and $\Bar{\hat{y}}$ is the average of the predictions $\hat{y}_b$, where $b \in [B]$. The aleatoric uncertainty is estimated from the residual sum of squares of the 10 latest predictions, given by
\begin{equation}
    \hat{\sigma}^2_a = \frac{1}{10} \sum_{i=1}^{10} (y_i - \hat{y}_i)^2 .
\end{equation}
Finally, the total uncertainty is calculated by 
\begin{equation}
    \hat{\sigma}^2 = \sqrt{\hat{\sigma}^2_{e} + \hat{\sigma}^2_a} ,
    \label{eq:total_uncertainty_MCNN}
\end{equation}
which effectively represents the standard error in Equation \ref{eq:pred_error}.

\section{Prediction Process}\label{sec:prediction_process}
As mentioned in Section \ref{introduction}, we use an online learning approach in this study, as it fits well with the sequential nature of the data, where new observations become available over time. The online learning approach enables continuous learning over time and avoids the need to retrain the model when new samples become available. However, a negative aspect of this approach is the lack of a standardized way of performing feature selection and hyperparameter optimization \cite{streaming_overview}. The authors of \cite{LOSING20181261} present two approaches for hyperparameter optimization. The first approach is called the \textit{offline setting}, where a complete training set is used to find the hyperparameters, much like in the traditional machine learning paradigm. The optimal hyperparameters are subsequently used to validate the models on a test set. The second approach, called the \textit{online setting}, saves the first 20\% (or the first 1000) observations in a buffer to perform hyperparameter tuning on. The second approach does, however, make the assumptions that the selection of these hyperparameters is optimal throughout the rest of the stream and that concept drifts do not occur \cite{streaming_overview}. 

In addition to online learning, this study also uses personalized models, i.e., there is one model instance per vehicle. This setup makes feature and hyperparameter selection more challenging, as the features and parameters yielding the best results might differ among vehicles. We simplify the selection of hyperparameters and features for each machine learning model by selecting the ones yielding the lowest average MAE for all cars. The selection method is similar to the first approach described in \cite{LOSING20181261}. That is, feature selection and hyperparameter tuning for a model are performed on a subset of the vehicles. The selected features and hyperparameters are then used on the remaining vehicles to evaluate the models. 

\subsection{Selecting Well-Behaving Vehicles}
The vehicles used in the feature and hyperparameter selection, and the vehicles used in the evaluation of the prediction models, are not selected completely at random. Analysis of the target variables indicates that the occurrence of recurrent vehicle usage varies greatly between different vehicles, potentially making the usage patterns of some cars more difficult to predict than that of others. Inspired by Goebel and Voß \cite{driving_behavior}, we want to focus on the vehicles that, despite the COVID-19 pandemic, have some regularity in the usage pattern. In their work, they predict the first daily departure time of commuter vehicles, and develop a selection method for finding “well-behaving” vehicles. The goal of this selection method is to find vehicles that regularly travel to work and have similar departure times each day. They do this by selecting vehicles with a low variance in the departure time and the number of weekdays where drives occur.

One problem with utilizing the same approach in this study is that it might favor cars with a similar departure time every day, and disregard cars whose departure times vary per day but are the same across weeks. To also capture the weekly patterns, we want to find cars where similar observations of departure times and trip distances are frequently observed. Therefore, we propose a different selection method, where the clustering tendency is measured, i.e., how well the data can be clustered. A strong clustering tendency indicates that similar observations are frequently observed and that a recurrent vehicle usage pattern exists.

The clustering tendency is measured by a statistical test called the Hopkins test \cite{CROSS1982315}. The test compares the joint distribution of the target variables with a two-dimensional uniform distribution by measuring distances between and within clusters. More specifically, for each sampled point in the joint distribution of target vectors, the Euclidean distance to its nearest neighbour is measured, as well as the distance to the nearest point in a uniform distribution of the same range as the target vectors.

We select the 100 cars with the highest clustering tendency. Out of those, 80\% are used in the feature selection and the hyperparameter selection, and 20\% are used in the evaluation of the prediction models. 

\subsection{Feature Selection}

The signals described in Section \ref{features} generate a large number of features. Reducing the number of features is often desired as it reduces the computational cost, and may generalize the model better. The feature selection process in this study consists of two steps. First, we perform a general feature selection to reduce the number of features, and the second step involves selecting the subset of features that produces the best result for each model.

The first step in the feature selection is to remove any features that do not seem to be helpful when predicting the target variables, and we perform two techniques to find them. First, we use the Pearson correlation coefficient to find features with no linear relationship to the target. To generalize over multiple cars, the correlations between the variables are computed for each vehicle separately and then averaged. The second technique is forward sequential feature selection, which starts from a prediction model considering only one feature, and then sequentially adds the features that improve the MAE the most, until a pre-determined number of features is reached.
For each target metric, linear regression and random forest are used to find the most significant features. In contrast to online learning, this step is performed using batch learning to train the models, as this seems to yield the most stable selection. For each target variable, features disfavored by the correlation technique and the sequential feature selection are removed, with one exception. If one feature of a one-hot encoded vector for a categorical variable is favored, all other features of the same variable are kept.

To find features that are highly correlated with each other, we look at their multicollinearity. High multicollinearity among the features might indicate that there are redundant features. To avoid this, we calculate a subset of the numeric features with low \acrfull{vif} \cite{multicollinearity} scores. The method is summarized in the following way:
\begin{enumerate}[leftmargin=*]
    \item \textit{Calculate the \acrshort{vif} for all numerical features.}
    \item \textit{If there exists features with a \acrshort{vif} greater than 10, remove the feature with the highest \acrshort{vif}.}
    \item \textit{Repeat step 1 and 2 until there are no features with a \acrshort{vif} higher than 10.}
\end{enumerate}

The second step in the feature selection is done by performing a backward sequential feature selection on each prediction model. Backward sequential feature selection is a greedy algorithm that attempts to find the subset of features that yields the best score, in our case the lowest aggregated \acrshort{mae} for all cars. Given a model and $n$ features, the algorithm first computes the model's \acrshort{mae} using all $n$ features by performing progressive validation, an evaluation method that fits well with how online models are used in practice and is introduced in \cite{blum1999beating}. Next, each feature is sequentially disregarded from the model, one at a time, and the model's \acrshort{mae} is computed for all subsets with $n-1$ features. We remove the feature which, by its removal, improves the model's \acrshort{mae} the most. This procedure of removing features continues until the model's \acrshort{mae} does not improve by removing any more features. {Using this procedure, for the task of driving distance prediction, out of a total of 60 generated features, 12 are selected for \acrshort{qr}, 51 for \acrshort{qknn}, 58 for \acrshort{qarf} and 58 for \acrshort{mcnn}. Similarly, out of a total of 55 features generated for departure time prediction, 13 are selected for \acrshort{qr}, 33 for \acrshort{qknn}, 52 for \acrshort{qarf} and 55 for \acrshort{mcnn}. Finally, we standardize each feature vector.}

\subsection{Hyperparameter Selection}
We perform hyperparameter selection by using \textit{Grid Search} \cite{GhawiPfeffer}. For each model, a subset of the hyperparameter space is chosen for evaluation. The Grid Search algorithm performs an exhaustive search among the possible combinations of hyperparameters. For each combination, the model is evaluated using progressive validation, and the combination which yields the lowest \acrshort{mae} over all cars is selected. 

\subsection{Evaluation}
The evaluation metrics used to evaluate the model performance are \gls{mae}, \gls{mape}, \gls{picp}, and \gls{mpiw}. \acrshort{picp} measures the percentage of observations correctly contained in the prediction interval, which is defined as
\begin{equation}
    PICP = \frac{1}{n} \sum_{i=1}^{n} c_i ,
\end{equation}
where $n$ is total number of observations, while $c_i = 1$ if the prediction is in the prediction interval and $0$ otherwise. \acrshort{mpiw} instead measures the breadth of the interval, which may be viewed as how certain the model is in its predictions. A narrow interval indicates high certainty. \acrshort{mpiw} is defined as 
\begin{equation}
    MPIW = \frac{1}{n} \sum_{i=1}^{n} ((\hat{y}+z_{\frac{\tau}{2}}\sigma) - (\hat{y}-z_{\frac{\tau}{2}}\sigma)) ,
\end{equation}
where $\hat{y}+z_{\frac{\tau}{2}}\sigma$ and $\hat{y}-z_{\frac{\tau}{2}}\sigma$ defines the upper and lower bound of the prediction interval respectively.

Each of the online prediction models {is implemented using Python with the libraries \textit{River}, \textit{PyTorch} and \textit{scikit-learn}.} They are evaluated using progressive validation on a held-out test set of cars, with the selected features and hyperparameters \cite{blum1999beating}. The performance of each type of model is calculated both on an aggregated level for all cars in the test set, and also for individual cars. To allow the prediction models a chance to learn before evaluation, we do not consider the predictions of the first 20 drives on each vehicle when calculating \acrshort{mae}, \acrshort{mape}, \acrshort{picp}, and \acrshort{mpiw}.

\section{Results and Discussion}\label{sec:results}
In this section, we describe the results of performing progressive validation on a held-out test set. First, we look at how well the online models could predict the departure time and distance of the first drive each day. Second, we look at how well the online models could estimate prediction intervals for the departure time and distance.

\subsection{Prediction Errors}
Table \ref{tab:td_covid} displays the result of predicting the distance of the first drive each day, {where the overall mean and standard deviation of the trip distances in the held-out test set are 24.73 km and 34.65 km, respectively.} All prediction models achieve a lower \acrshort{mae} than the baseline model, however the models are only slightly more accurate. The best performing prediction model, \acrshort{qarf}, has a MAE over all cars of 13.37 km, thus it is, on average, only around 3.5 km more accurate than just taking the mean of all previous driving distances. However, by looking at the \acrshort{mape}, it is clear that some of the models yield a significant improvement. The baseline has a \acrshort{mape} of 190 when considering all cars, while \acrshort{qknn} has 105. It is expected that the \acrshort{mape} is high, as a large proportion of the drives are short. Looking at the percentage of driving distances which are predicted within 5 km of the true distance also shows that some models yield a significant improvement. When considering all cars, 21\% of the baseline's driving distance predictions are less than 5 km off. The best models manage to predict more than half of the drives within a 5 km interval of the true distance.

\begin{table}
\centering
\caption{Model performance when predicting driving distance}
\label{tab:td_covid}
\begin{tabular}{|l|l|l|l|}
\hline
\multicolumn{1}{|c|}{Model} & \multicolumn{1}{c|}{MAE (km)}             & \multicolumn{1}{c|}{MAPE}                & \multicolumn{1}{l|}{\begin{tabular}[c]{@{}l@{}} \% with error \\ $\leq$ 5 km\end{tabular}} \\ \cline{1-4} 
Mean                                         & \multicolumn{1}{l|}{16.95}  & \multicolumn{1}{l|}{190.31}  & \multicolumn{1}{l|}{21}      \\ \hline
QR                                           & \multicolumn{1}{l|}{14.24}  & \multicolumn{1}{l|}{131.21}  & \multicolumn{1}{l|}{40}      \\ \hline
QKNN                                         & \multicolumn{1}{l|}{13.52}  & \multicolumn{1}{l|}{104.97}  & \multicolumn{1}{l|}{52}      \\ \hline
QARF                                         & \multicolumn{1}{l|}{13.37}  & \multicolumn{1}{l|}{107.28}  & \multicolumn{1}{l|}{52}      \\ \hline
MCNN                                         & \multicolumn{1}{l|}{15.68}  & \multicolumn{1}{l|}{161.96}  & \multicolumn{1}{l|}{28}      \\ \hline
\end{tabular}
\end{table}

\begin{figure}
    \centering
    \includegraphics[width=0.4\textwidth]{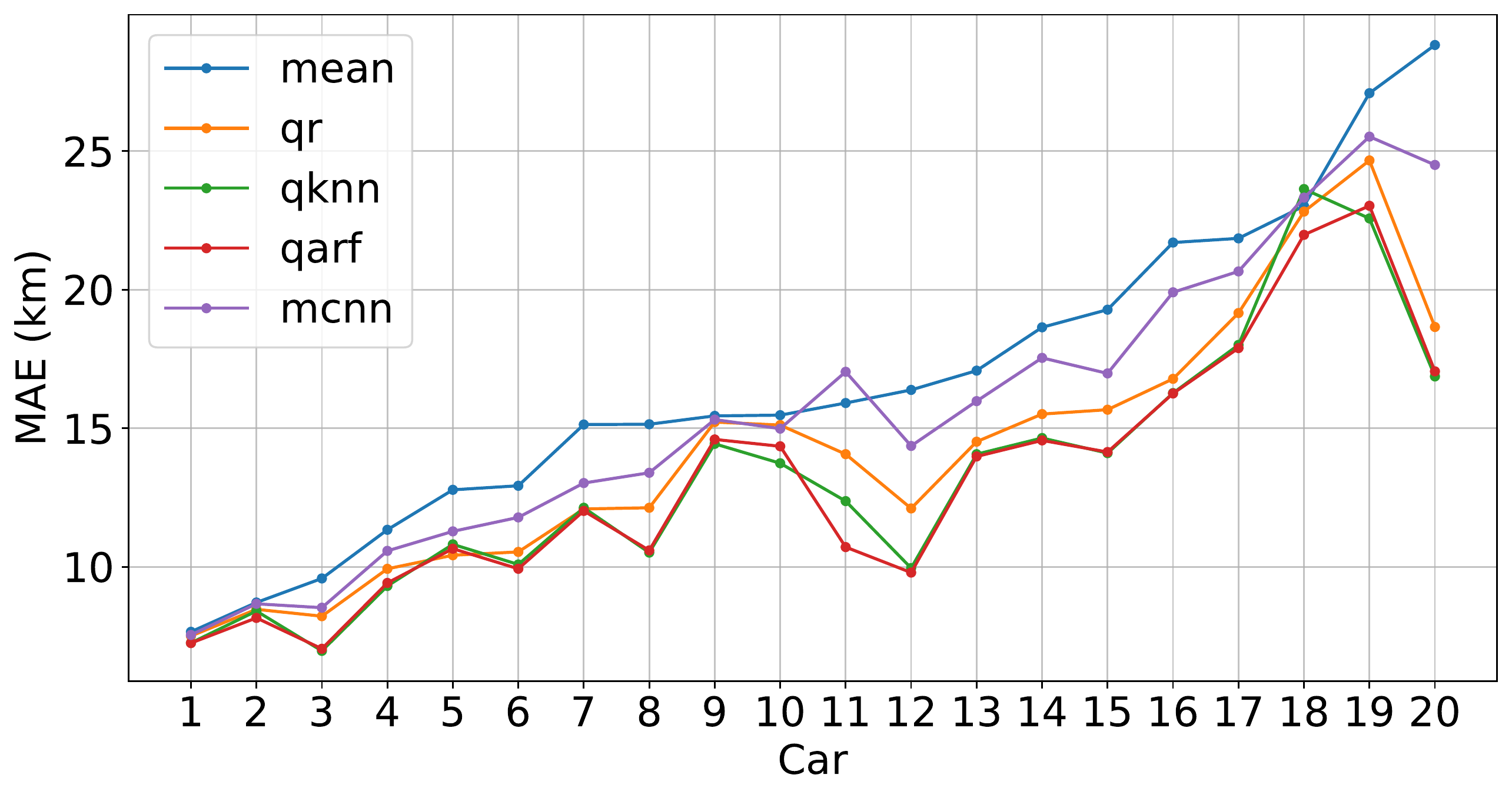}
    \caption{The \acrshort{mae} per vehicle when predicting driving distance. The cars are sorted by the \acrshort{mae} of the baseline (mean)}
    \label{fig:mae_td_per_car}
\end{figure}

A more exhaustive view of the models' performances can be observed in Figure \ref{fig:mae_td_per_car}, where the \acrshort{mae} for each vehicle in the held-out test set is illustrated. It is clear that the models' ability to predict vehicle usage differs considerably between vehicles. Furthermore, we observe that the models produce significantly more accurate predictions than the baseline on some vehicles, while for other vehicles there is no improvement. 

The models' performance when predicting the first departure time of each day is displayed in Table \ref{tab:ttl_covid}, {with an overall mean departure time in the test set of 10.72 hours and a standard deviation of 4.12 hours.} As in the driving distance case, all models achieve a lower \acrshort{mae} than the baseline, however, the improvement is rather slight. \acrshort{qr} achieves the lowest \acrshort{mae} of 2.75 hours, which makes it on average around 30 min more accurate than the baseline. However, looking at the percentage of predicted departure times that are within an hour of the true departure time, it is clear that most of the prediction models are fairly more accurate than the baseline. The baseline only predicts 14\% of the departure times within an hour of the actual departure, compared to 34\% for the best performing prediction models. When looking at the \acrshort{mape}, we can see a significant difference compared to the trip distance prediction. This makes sense when considering the nature of the predictions. In the departure time setting, it is reasonable to assume that a trip is performed a couple of hours into the day, which is considerably larger than the \acrshort{mae} displayed in Table \ref{tab:ttl_covid}. In the distance setting, it is likely that the distance driven might be short. Depending on the actual driven distance, it may happen that the models predict double the distance, even if it is just a matter of a few kilometers.

\begin{table}
\centering
\caption{Model performance when predicting departure time.}
\label{tab:ttl_covid}
\begin{tabular}{|l|l|l|l|}
\hline
\multicolumn{1}{|c|}{Model} & \multicolumn{1}{c|}{MAE (h)}             & \multicolumn{1}{c|}{MAPE}                & \multicolumn{1}{l|}{\begin{tabular}[c]{@{}l@{}}\% with error \\ $\leq$ 60 min\end{tabular}} \\ \cline{1-4} 
Mean                                         & \multicolumn{1}{l|}{3.22}   & \multicolumn{1}{l|}{82.89}   & \multicolumn{1}{l|}{14}         \\ \hline
QR                                           & \multicolumn{1}{l|}{2.75}   & \multicolumn{1}{l|}{76.40}   & \multicolumn{1}{l|}{34}         \\ \hline
QKNN                                         & \multicolumn{1}{l|}{2.94}   & \multicolumn{1}{l|}{72.47}   & \multicolumn{1}{l|}{34}         \\ \hline
QARF                                         & \multicolumn{1}{l|}{2.90}   & \multicolumn{1}{l|}{71.02}   & \multicolumn{1}{l|}{32}         \\ \hline
MCNN                                         & \multicolumn{1}{l|}{3.03}   & \multicolumn{1}{l|}{79.34}   & \multicolumn{1}{l|}{18}         \\ \hline
\end{tabular}
\end{table}

\begin{figure}
    \centering
    \includegraphics[width=0.4\textwidth]{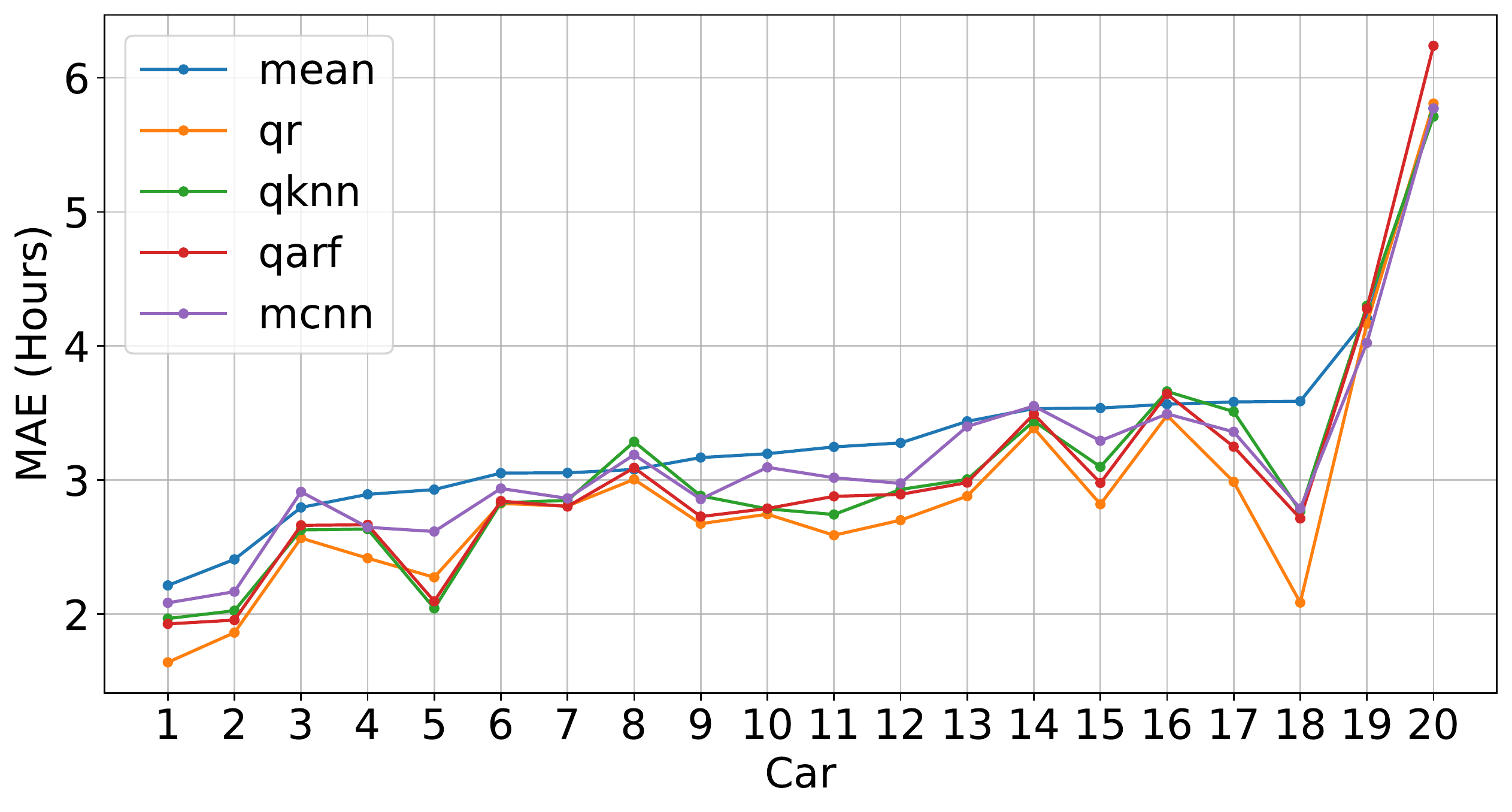}
    \caption{The \acrshort{mae} per vehicle when predicting departure time. The cars are sorted by the \acrshort{mae} of the baseline (mean)}
    \label{fig:mae_ttl_per_car}
\end{figure}

Figure \ref{fig:mae_ttl_per_car} shows the \acrshort{mae} of each vehicle in the test set, giving a more extensive view of the models' performances. As in the case of driving distance, the figure illustrates that the models' accuracies differ considerably among vehicles. We may again note that the models produce significantly more accurate predictions than the baseline on some vehicles, while for other vehicles there are no accuracy improvements.

The evaluation implies consistency across the two prediction targets in the models' performances. The least accurate model in almost all cases is the baseline. \acrshort{mcnn} slightly outperforms the baseline in most cases, but is less accurate than \acrshort{qr}, \acrshort{qknn}, and \acrshort{qarf}. \acrshort{qr} performs well when predicting departure times, yielding the lowest \acrshort{mae} when considering all cars. However, \acrshort{qr} does not perform as well when predicting driving distance and is less accurate than \acrshort{qknn} and \acrshort{qarf} in this case. Figures \ref{fig:mae_td_per_car} and \ref{fig:mae_ttl_per_car} show that \acrshort{qknn} and \acrshort{qarf} most often are among the top-performing models. 

\begin{figure}%
    \centering
    \captionsetup[subfloat]{font=scriptsize,labelfont=scriptsize}
    \subfloat[\centering Driving distance \label{fig:over_time_td_midnight_mae}] {\includegraphics[width=0.24\textwidth]{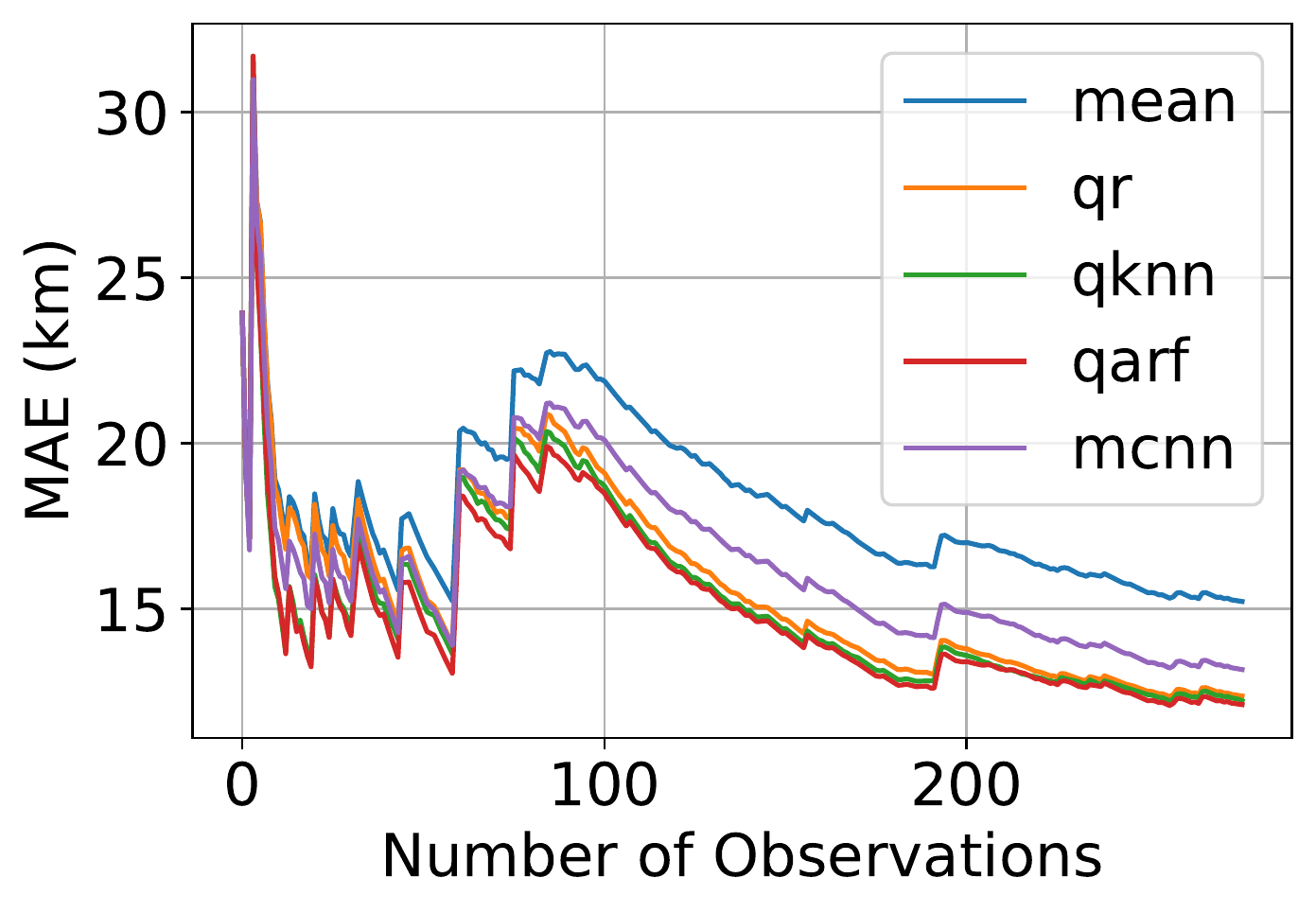}}
    \subfloat[\centering Departure time \label{fig:over_time_ttl_midnight_mae}] {\includegraphics[width=0.24\textwidth]{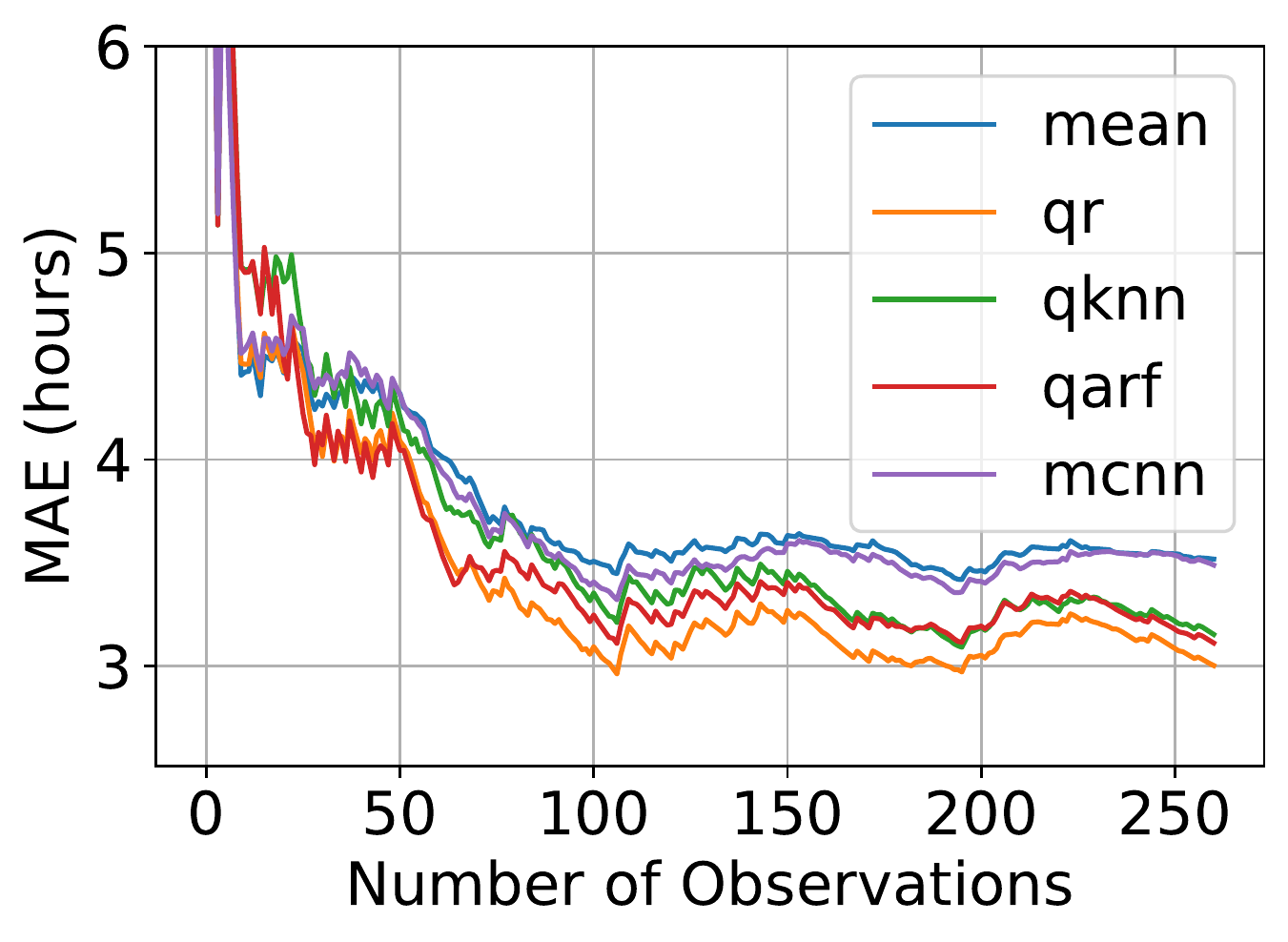}}
    \caption{Examples from a single vehicle on how the \acrshort{mae} changes as more drives are observed when predicting driving distance and departure time.}
    \label{fig:over_time_mae_midnight}
\end{figure}

Figure \ref{fig:over_time_mae_midnight} shows examples from a single vehicle of how \acrshort{mae} changes when the number of observations increases. We can note that it is not always the case that more observations yield a lower \acrshort{mae}. Periodically, the \acrshort{mae} might increase due to changing vehicle usage. For example, the summer period seems to often cause temporary changes in vehicle usage, resulting in increased prediction errors.

From Figure \ref{fig:over_time_mae_midnight} we can also note that all models, including the baseline, have similar curves and that the models' \acrshort{mae}s seem to be affected in similar ways when observing new data. This suggests that all models are making quite similar predictions. One reason for this may be that no model manages to learn the full vehicle behavior, but rather they all learn some average or the most common drive. 

\subsection{Uncertainty}

Table \ref{tab:uncertainty} displays how well the models estimate 90\% prediction intervals for driving distances and departure times. Both tables show that the models generally do not reach the desired coverage probability. However, \acrshort{mcnn}, \acrshort{qarf} and \acrshort{qknn} all have coverage probabilities close to 90\%. \acrshort{mpiw} is very large in all these cases, spanning tens of kilometers or several hours. \acrshort{qr} produces much narrower intervals, but the \acrshort{picp} is significantly less than desired.

\begin{table}
\centering
    \caption{Model performance when creating prediction intervals}
    \label{tab:uncertainty}
\begin{tabular}{|l|ll|ll|}
\hline
\multicolumn{1}{|c|}{\multirow{2}{*}{Model}} & \multicolumn{2}{l|}{Driving Distance} & \multicolumn{2}{l|}{Departure Time} \\ \cline{2-5} 
\multicolumn{1}{|c|}{}                       & \multicolumn{1}{l|}{PICP}   & MPIW    & \multicolumn{1}{l|}{PICP}  & MPIW   \\ \hline
QR                                           & \multicolumn{1}{l|}{0.56}   & 26.22   & \multicolumn{1}{l|}{0.70}  & 9.03   \\ \hline
QKNN                                         & \multicolumn{1}{l|}{0.81}   & 61.45   & \multicolumn{1}{l|}{0.83}  & 10.83  \\ \hline
QARF                                         & \multicolumn{1}{l|}{0.84}   & 79.83   & \multicolumn{1}{l|}{0.85}  & 11.37  \\ \hline
MCNN                                         & \multicolumn{1}{l|}{0.92}   & 87.31   & \multicolumn{1}{l|}{0.88}  & 13.23  \\ \hline
\end{tabular}
\end{table}

The results show that the prediction intervals are not perfectly accurate, as they often do not achieve the desired coverage probability. However, this is somewhat expected. As mentioned in Section \ref{sec:prediction_process}, we begin measuring the \acrshort{picp} after the models have observed 20 drives. It is unlikely that the first 20 drives represent a vehicle's full usage during a year, and the models may therefore not be expected to quantify the uncertainty perfectly yet. Instead, the models continue to improve as more drives are observed.

\begin{figure}%
    \centering
    
    \captionsetup[subfloat]{font=scriptsize,labelfont=scriptsize}
    
    \subfloat[\centering \acrshort{picp} Driving distance\label{fig:over_time_td_midnight_picp}] {\includegraphics[width=0.24\textwidth]{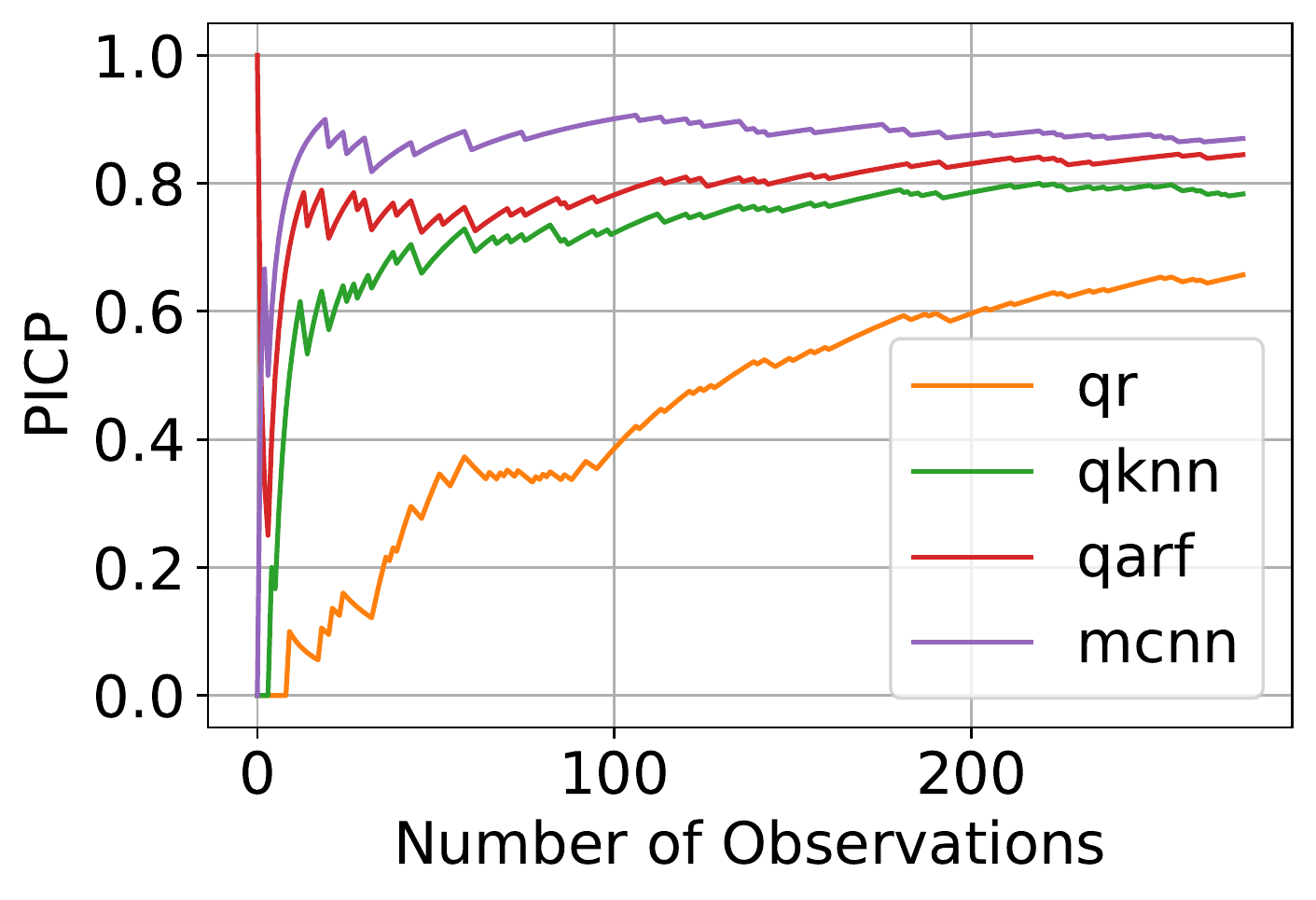}}
    \subfloat[\centering \acrshort{mpiw} Driving distance \label{fig:over_time_td_midnight_mpiw}] {\includegraphics[width=0.24\textwidth]{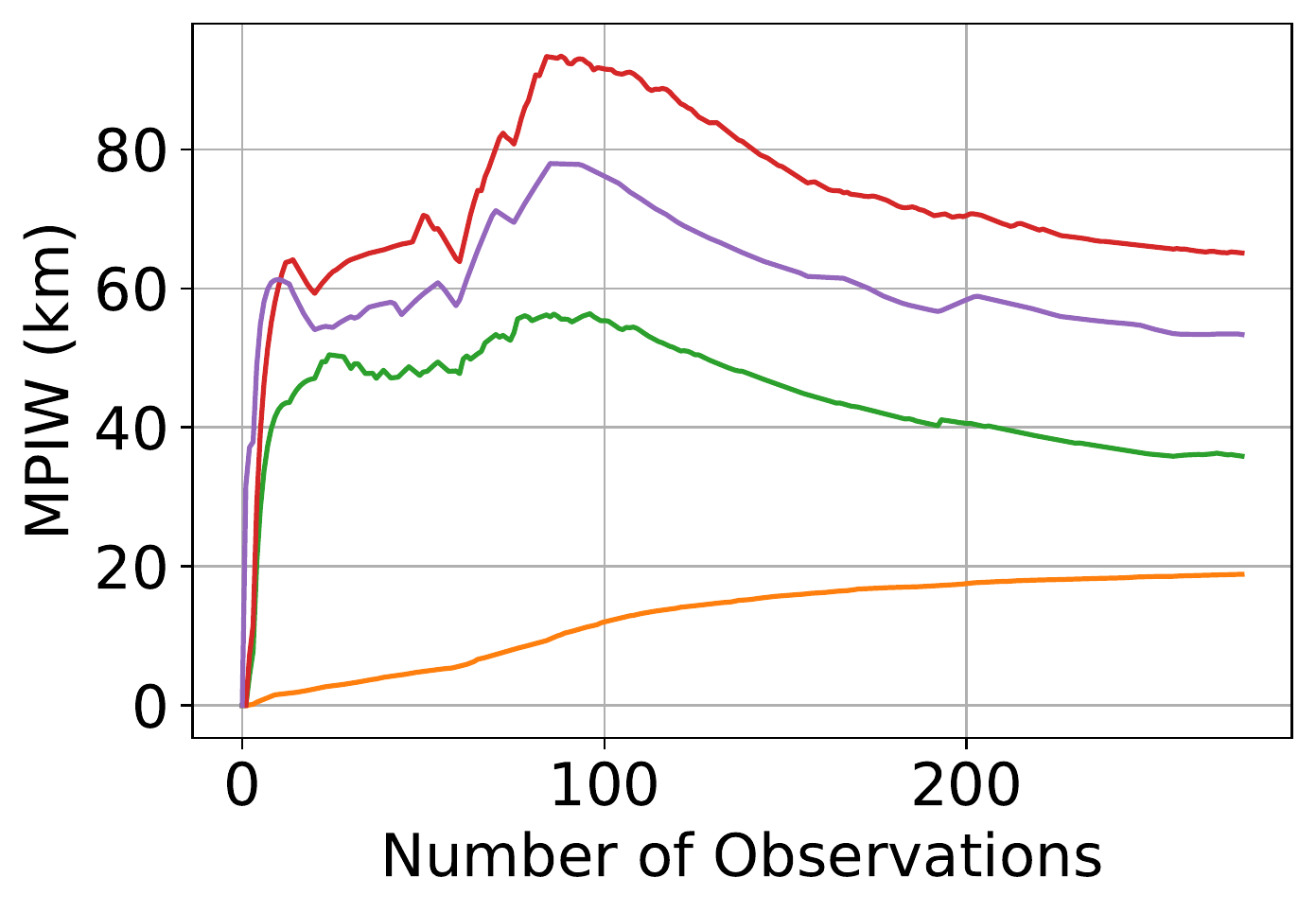}}
    \\
    \subfloat[\centering \acrshort{picp} Departure time \label{fig:over_time_ttl_midnight_picp}]
    {\includegraphics[width=0.24\textwidth]{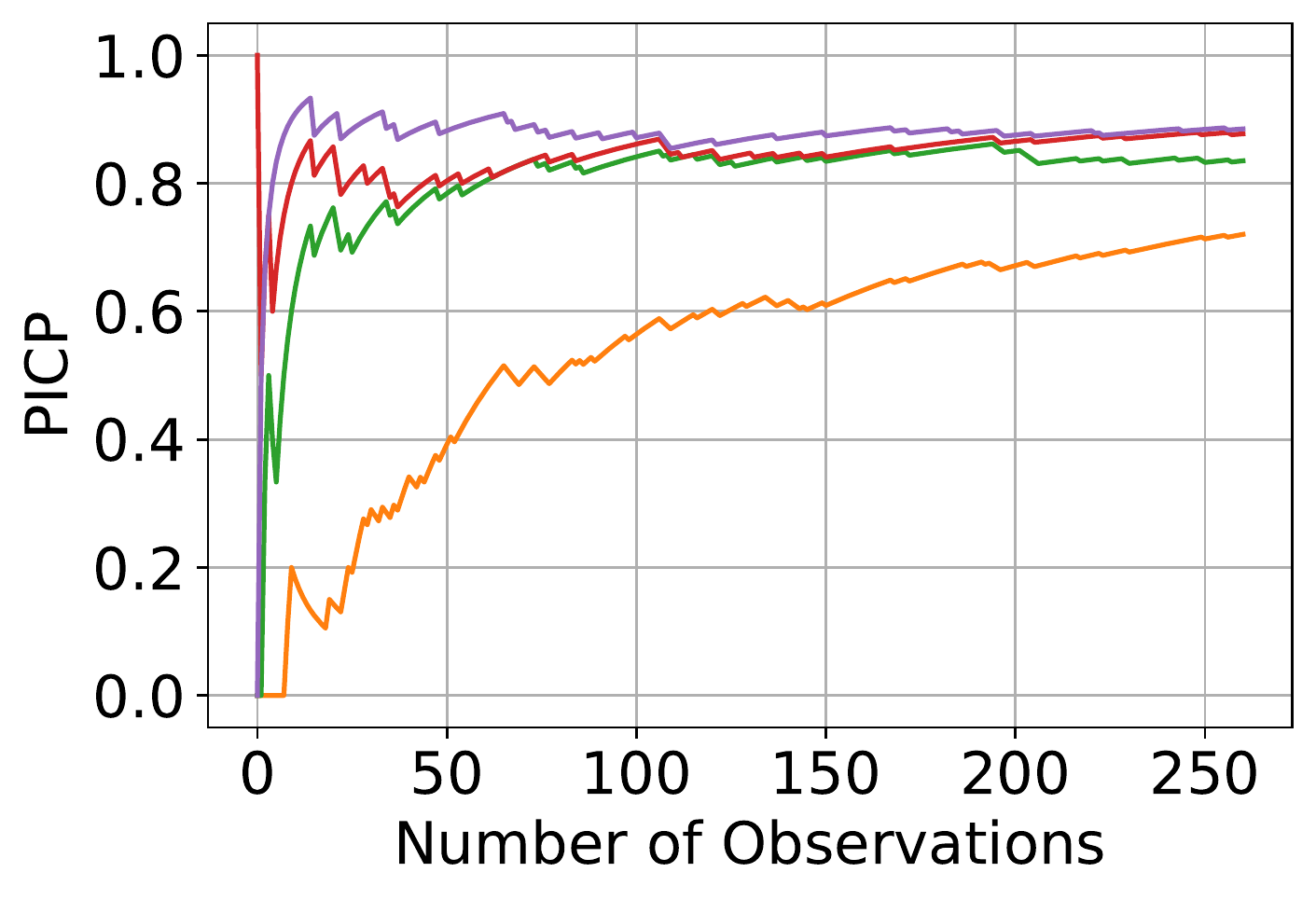}}
    \subfloat[\centering \acrshort{mpiw} Departure time \label{fig:over_time_ttl_midnight_mpiw}] {\includegraphics[width=0.24\textwidth]{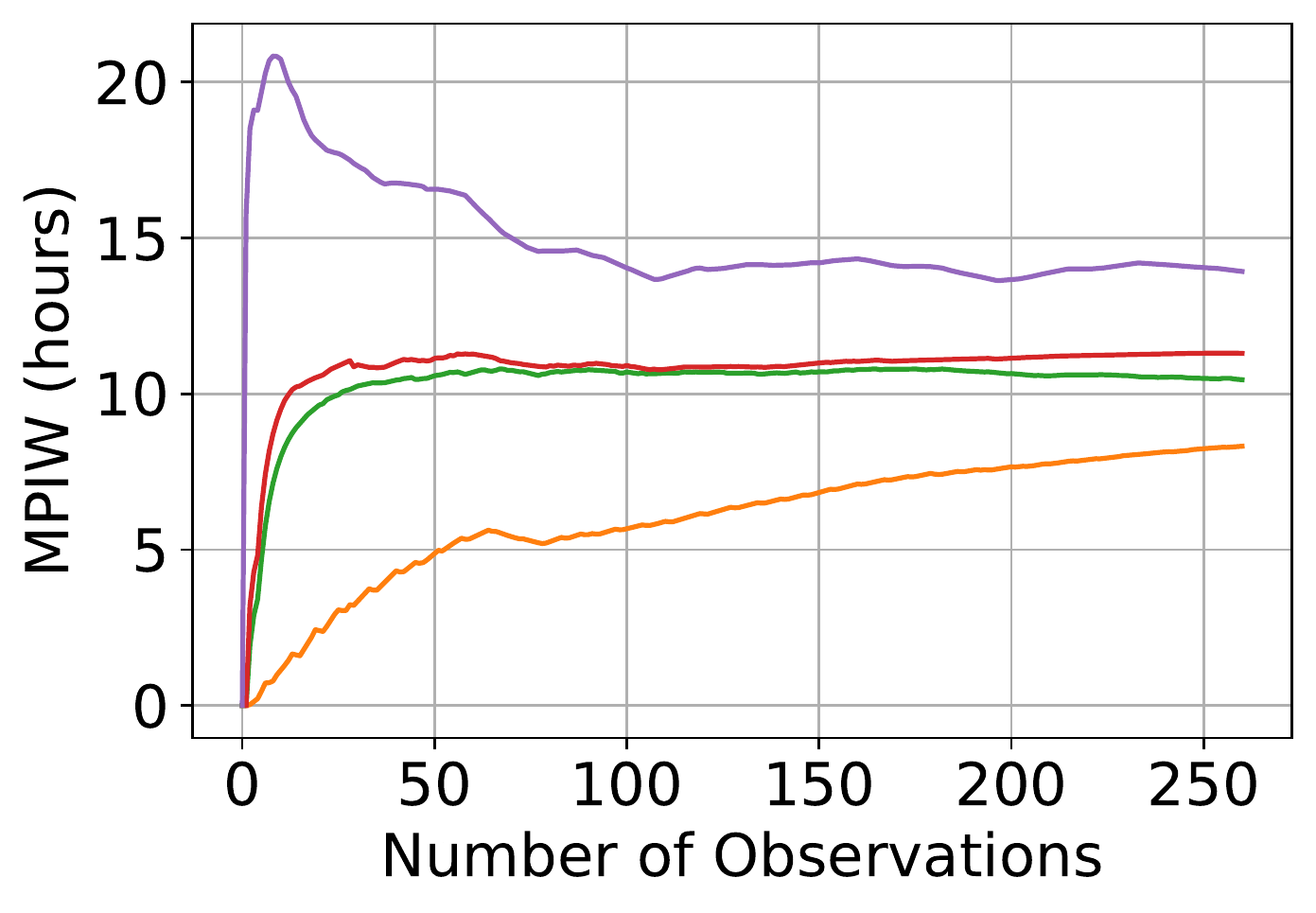}}
    \caption{Examples from a single vehicle on how the \acrshort{picp} and \acrshort{mpiw} change over time when predicting driving distance and departure time.}
    \label{fig:over_time_td_midnight_picp_mpiw}
    
\end{figure}

Figure \ref{fig:over_time_td_midnight_picp_mpiw} illustrates, for a single vehicle, how \acrshort{picp} and \acrshort{mpiw} change as more observations are received. Both figures show how \acrshort{qknn}, \acrshort{qarf}, and \acrshort{mcnn} quickly reach a fairly high \acrshort{picp}, while \acrshort{qr} needs many more observations to begin approaching a comparable prediction coverage. The pattern is similar in the case of \acrshort{mpiw}. The prediction intervals of \acrshort{qknn}, \acrshort{qarf}, and \acrshort{mcnn} have a large width after only a few observations, while \acrshort{qr} needs many more observations to reach a similar \acrshort{mpiw}.

\section{Conclusion}\label{sec:conclusion}

In this study, we predict the departure time and driving distance of the first drive of the day during COVID-19, using various online learning models. 
The prediction models are evaluated according to their error rate and how well they could estimate a 90\% prediction interval. 
When we look at all drives of all cars in the test set, the proposed prediction models generally yield an \acrshort{mae} of slightly below 3 hours when predicting departure time and 13-16 km when predicting driving distance. 
In the case of departure time, \acrshort{qr} displays the best predictive performance while \acrshort{qarf} and \acrshort{qknn} perform best in the driving distance case. The models' ability to quantify uncertainty varies, with \acrshort{mcnn} and \acrshort{qarf} generally estimating the most accurate prediction intervals. The results indicate the surprising difficulty of predicting vehicle usage.

\section*{Acknowledgements}
Part of this study was performed by T.~Lindroth and A.~Svensson as thesis project \cite{lindroth2022predicting} for the degree of M.Sc., with the support of Volvo Car Corporation, which also provided data and resources for the study. 

\bibliographystyle{IEEEtran}
\bibliography{mendeley}
\vspace{-10mm}
\begin{IEEEbiography}
[{\includegraphics[width=1in, height=1.25in,clip, keepaspectratio]{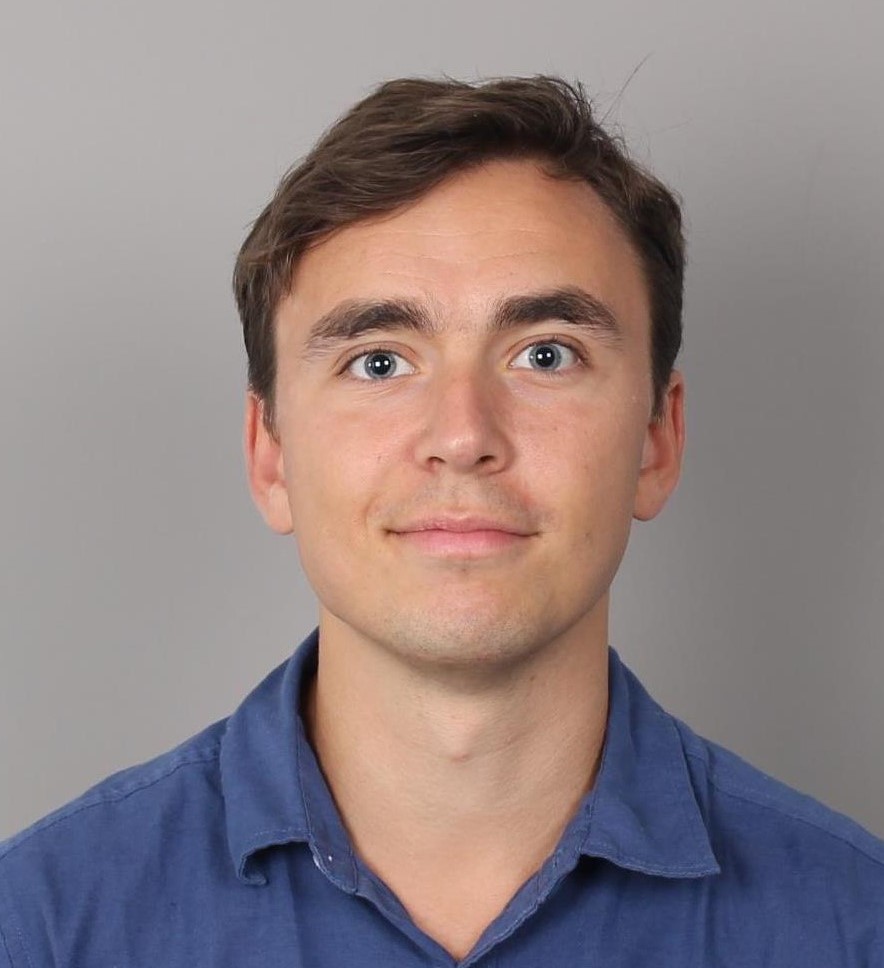}}]{Tobias Lindroth} received the B.Sc degree in Software Engineering in 2020 and the M.Sc. degree in Data Science and AI in 2022, from Chalmers University of Technology, Gothenburg, Sweden. His M.Sc. thesis is in the field of incremental machine learning applied on vehicle usage prediction to improve the energy efficiency for battery electric vehicles. He started working at Volvo Car Corporation in 2022 as a system design engineer within energy efficiency.
\end{IEEEbiography}
\vspace{-10mm}
\begin{IEEEbiography}
    [{\includegraphics[width=1in,height=1.25in,clip,keepaspectratio]{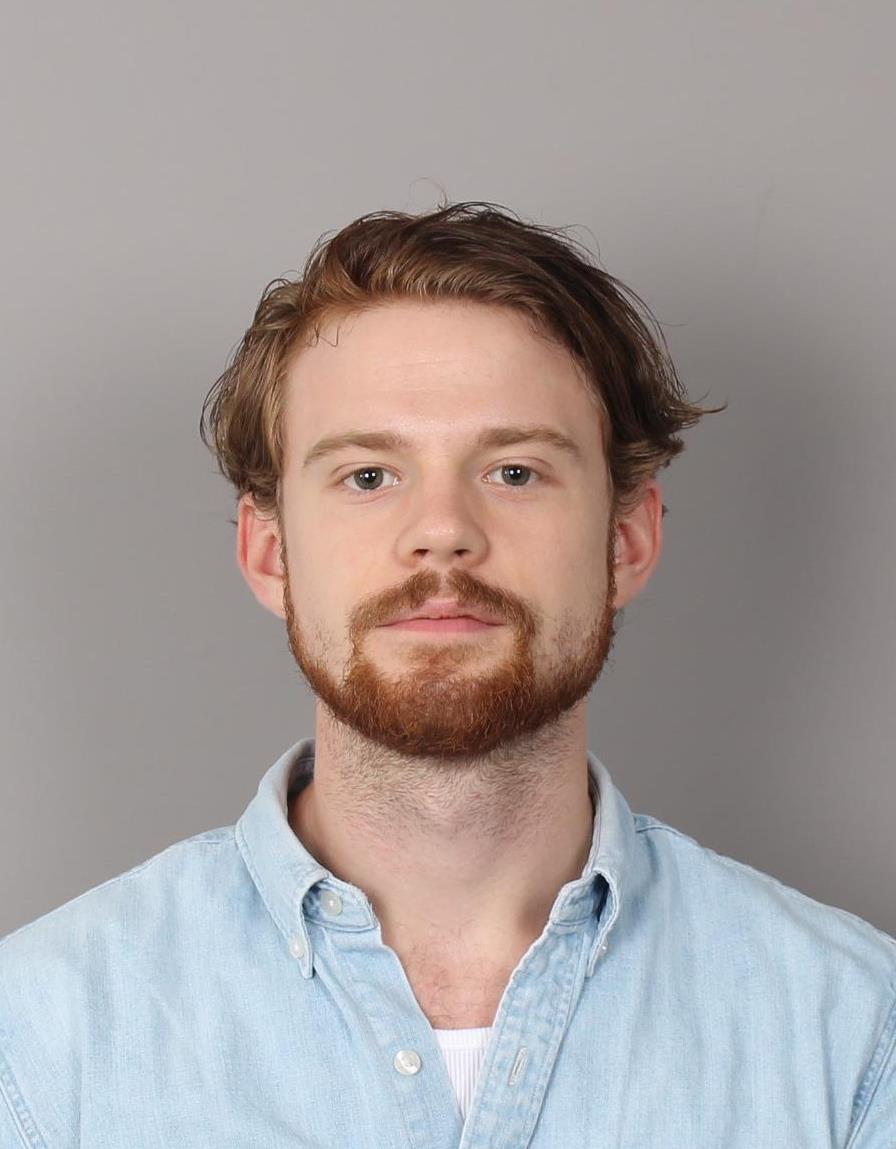}}]{Axel Svensson}
received the B.Sc degree in Software Engineering in 2020 and the M.Sc. degree in Data Science and AI in 2022, from Chalmers University of Technology, Gothenburg, Sweden. His M.Sc. thesis is in the field incremental machine learning to improve the energy efficiency for battery electric vehicles. He started working at Volvo Car Corporation in 2022 as a system design engineer within energy efficiency. 
\end{IEEEbiography}
\vspace{-10mm}
\begin{IEEEbiography}
    [{\includegraphics[width=1in,height=1.25in,clip,keepaspectratio]{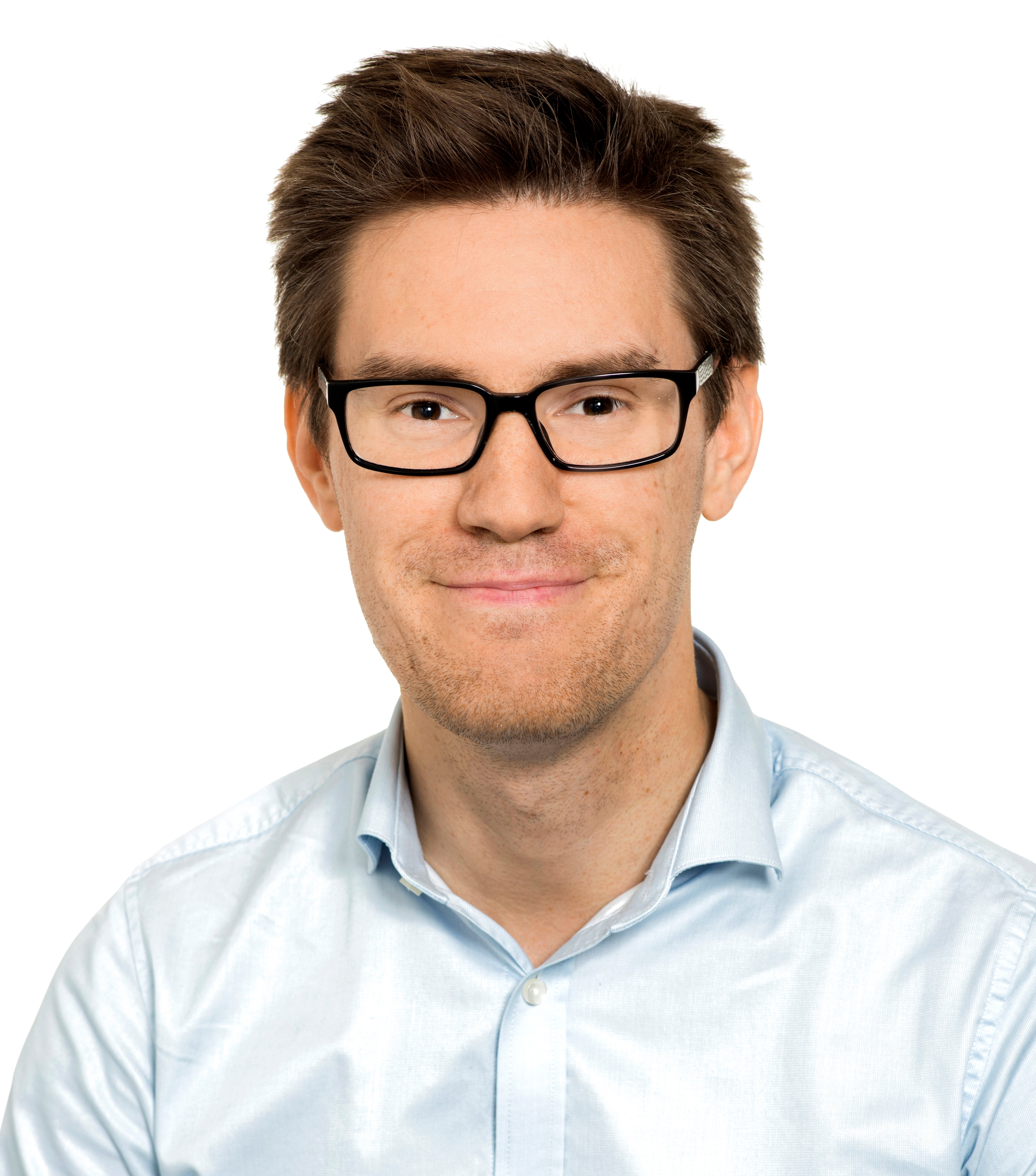}}]{Niklas Åkerblom}
received the M.Sc. degree in computer science and engineering in 2013 from Chalmers University of Technology, and is currently pursuing a Ph.D. in data science and AI at the same university. He started working as a software engineer at Volvo Car Corporation in 2013 and switched to an industrial Ph.D. position within the same company in 2019. His research interests include machine learning, data science, sequential decision making under uncertainty, and navigation algorithms.
\end{IEEEbiography}
\vspace{-10mm}
\begin{IEEEbiography}
    [{\includegraphics[width=1in,height=1.25in,clip,keepaspectratio]{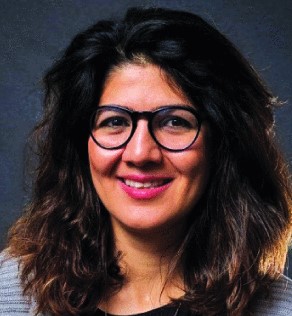}}]{Mitra Pourabdollah}
received her B.Eng. degree in Electrical Engineering from Shiraz University, Shiraz, Iran in 2006, M.Sc. degree in Systems, Control, and Robotics from Royal Institute of Technology, Stockholm, Sweden in 2009, and Ph.D. degree in Automatic Control from Chalmers University of Technology, Gothenburg, Sweden, in 2015. She joined Volvo Car Corporation as researcher in 2015. Her main focuses are the application of energy optimization and control in automotive and traffic area.
\end{IEEEbiography}
\vspace{-10mm}
\begin{IEEEbiography}
    [{\includegraphics[width=1in,height=1.25in,clip,keepaspectratio]{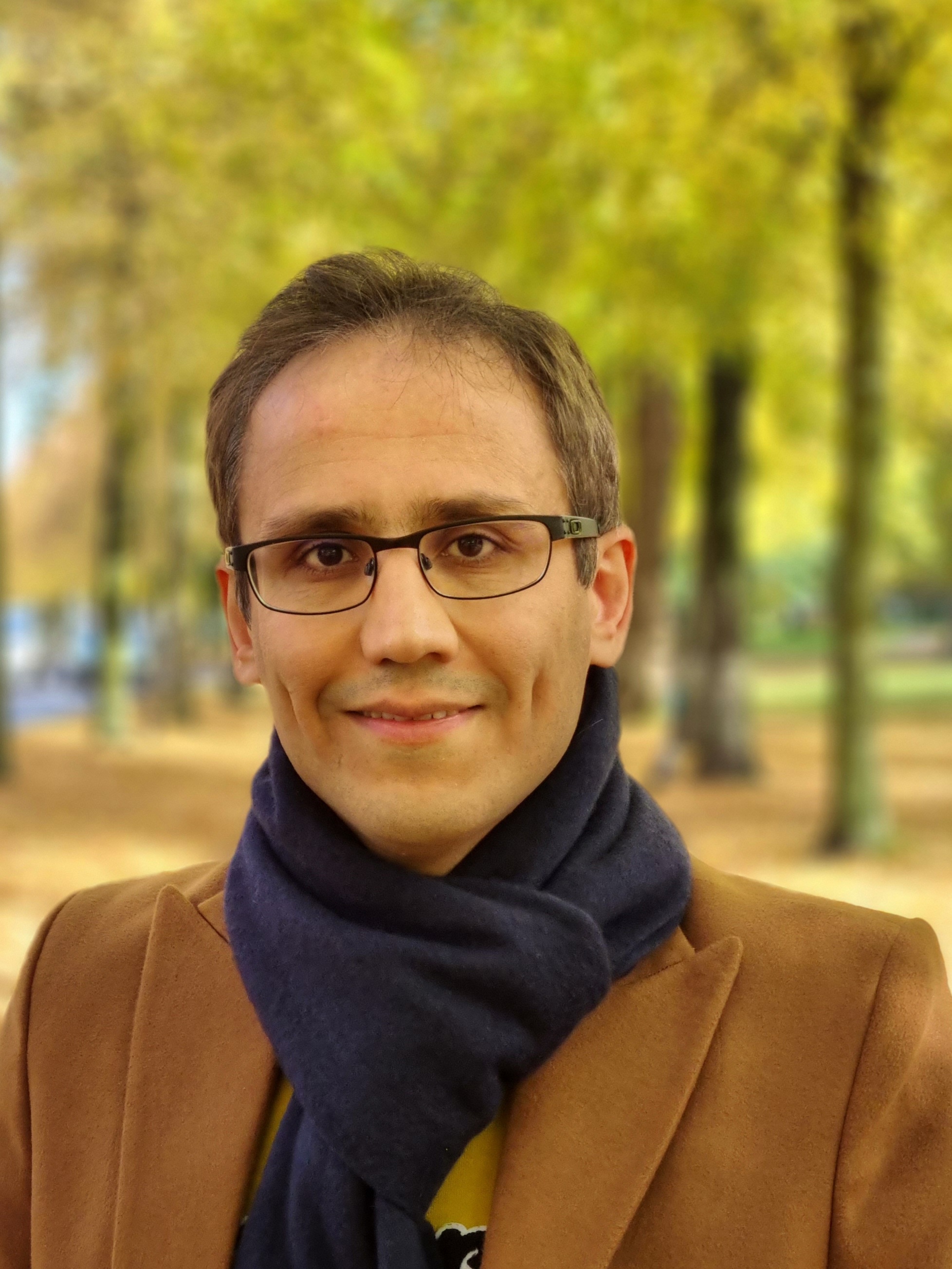}}]{Morteza Haghir Chehraghani}
is an associate professor in machine learning, artificial intelligence and data science at the Department of Computer Science and Engineering, Chalmers University of Technology. He holds a Ph.D. from ETH Zurich in Computer Science, Machine Learning. After the Ph.D., he spent about four years at Xerox Research Centre Europe  as a researcher (Staff Research Scientist I, and then Staff Research Scientist II). He joined Chalmers in 2018 as an associate professor. 
\end{IEEEbiography}
\end{document}